\documentclass{article}

    \usepackage[preprint,nonatbib]{neurips_2022}

\usepackage[utf8]{inputenc} 
\usepackage[T1]{fontenc}    
\usepackage{hyperref}       
\usepackage{url}            
\usepackage{booktabs}       
\usepackage{amsfonts}       
\usepackage{nicefrac}       
\usepackage{microtype}      
\usepackage{xcolor}         

\usepackage{enumitem} 
\usepackage{overpic} 
\usepackage{color}
\usepackage{times}
\usepackage{epsfig}
\usepackage{algpseudocode}
\usepackage{multirow}
\usepackage{mathtools}
\usepackage{changepage}
\usepackage{color, colortbl}
\usepackage{arydshln}  
\usepackage{enumitem}  
\usepackage{algorithm}
\usepackage{bbm}
\usepackage{array}
\usepackage[normalem]{ulem}

\usepackage{placeins} 

\definecolor{turquoise}{cmyk}{0.65,0,0.1,0.3}
\definecolor{purple}{rgb}{0.65,0,0.65}
\definecolor{dark_green}{rgb}{0, 0.5, 0}
\definecolor{orange}{rgb}{0.8, 0.6, 0.2}
\definecolor{red}{rgb}{0.8, 0.2, 0.2}
\definecolor{darkred}{rgb}{0.6, 0.1, 0.05}
\definecolor{blueish}{rgb}{0.0, 0.3, .6}
\definecolor{light_gray}{rgb}{0.7, 0.7, .7}
\definecolor{pink}{rgb}{1, 0, 1}
\definecolor{greyblue}{rgb}{0.25, 0.25, 1}

\hypersetup{urlcolor=blue}
\makeatletter
\DeclareUrlCommand\ULurl@@{%
  \def\UrlLeft{\uline\bgroup}%
  \def\UrlRight{\egroup}}
\def\ULurl@#1{\hyper@linkurl{\ULurl@@{#1}}{#1}}
\DeclareRobustCommand*\ULurl{\hyper@normalise\ULurl@}
\makeatother

\newcommand\eqdef{\stackrel{\text{def}}{=}}

\definecolor{Gray}{gray}{0.95}

\usepackage{blindtext}

\renewcommand{\paragraph}[1]{\vspace{1em}\noindent\textbf{#1}.}
\begin{document}
\title{Membership Inference Attack Using\\ Self Influence Functions}

%

\author{%
  Gilad Cohen \\
  School of Electrical Engineering \\
  Tel Aviv University \\
  Tel Aviv, 69978 \\
  \texttt{giladco1@post.tau.ac.il} \\
  \And
  Raja Giryes \\
  School of Electrical Engineering \\
  Tel Aviv University \\
  Tel Aviv, 69978 \\
  \texttt{raja@tauex.tau.ac.il} \\
}

\maketitle
\begin{abstract}
Member inference (MI) attacks aim to determine if a specific data sample was used to train a machine learning model. Thus, MI is a major privacy threat to models trained on private sensitive data, such as medical records.
In MI attacks one may consider the black-box settings, where the model's parameters and activations are hidden from the adversary, or the white-box case where they are available to the attacker.
In this work, we focus on the latter and present a novel MI attack for it that employs influence functions, or more specifically the samples' self-influence scores, to perform the MI prediction. We evaluate our attack on CIFAR-10, CIFAR-100, and Tiny ImageNet datasets, using versatile architectures such as AlexNet, ResNet, and DenseNet. Our attack method achieves new state-of-the-art results for both training with and without data augmentations. Code is available at \ULurl{https://github.com/giladcohen/sif_mi_attack}.
\end{abstract}
\section{Introduction}
\label{Introduction}
Machine learning (ML) algorithms have advanced tremendously over the past decade and have been commonly used for a variety of tasks, including privacy sensitive application, such as medical imaging \cite{LUNDERVOLD2019102,CALLI2021102125}, conversations \cite{Devlin2019BERTPO}, face recognition \cite{Taigman2014DeepFaceCT}, and financial information \cite{Damrongsakmethee2017DataMA}. Most of these models are trained using sensitive user data which can be leaked later by an adversary from the models' parameters \cite{Shokri2017MembershipIA}.

Membership inference (MI) attacks aim to infer whether or not a specific sample was used to train a target ML model. This information can be detrimental if it falls to the wrong hands. For example, consider an ML model trained on blood tests of HIV patients, for predicting their reaction to a corona vaccine. If an adversary somehow obtains a patient's medical record, she can only observe the patient's blood reading and query the model for the predicted reaction, but she cannot deduce if the patient has HIV. However, if the adversary infers that the record was used to train the model, then she would know the patient has HIV. If this adversary is a health insurance company, it might increase the patient's insurance premium.

Many MI attacks make use of the class probability vector (or logits) at the output of the target model \cite{Li2020MembershipIA,Salem2019MLLeaksMA,Shokri2017MembershipIA,Yeom2018PrivacyRI}, since deep neural networks (DNNs) often tend to exhibit over-confidence for samples from their training set \cite{Rezaei2021OnTD}, a phenomenon that is largely attributed to overfitting \cite{Yeom2018PrivacyRI}. More recent studies do not assume access to model probability vectors and still achieve state-of-the-art (SOTA) MI accuracy by relying on the final predicted labels at the model output \cite{ChoquetteChoo2021LabelOnlyMI,DBLP:journals/corr/abs-2007-15528}.

MI attacks can operate under two threat model settings: \textit{white-box} or \textit{black-box}. The white-box setting assumes that the adversary has full information on the target model's architecture, parameters, activations, training process, and training data distribution. On the other hand, the black-box setting is more restrictive, allowing the adversary access only to the target model's input and outputs. All the aforementioned MI attacks use the black-box setting. Other works assumed white-box setting and tried to exploit other information from the target model \cite{Sablayrolles2019WhiteboxVB,leino2020stolen,Rezaei2021OnTD}, however their white-box methods could not achieve a significant improvement in the MI prediction accuracy compared to black-box attacks.

\textbf{Contribution.} In this work we introduce a novel white-box MI attack that can be applied to any ML model. The core idea of our attack model is that training samples have a direct influence on the loss of test samples, but not vice versa. For quantifying this effect, we use influence functions \cite{Koh17Understanding}, which determines how data points in the training set influence the target model's prediction for a given test sample. This measure quantifies how much a small upweighting of a specific training point in the target model's empirical error affects the loss of a test point. To speed up computation time, we utilize the self-influence function of a sample point on it own loss.

Given a sample point, we calculate its self-influence function (SIF\footnote{SIF refers both to the self-influence function score and the attack model that is based on it interchangeably, depending on the context.}) score, and query the target model for its label prediction. These two values alone are sufficient to infer if the sample belongs to the training set. Our attack model makes use of only two parameters and thus exhibits fast inference time. We evaluate our MI attack on several datasets trained on various target models with different architectures, showing its advantage over current SOTA attacks. Moreover, we also consider the MI defense of training with data augmentations, which is a common practice in neural network training, and present an adaptive attack model that negates it. Specifically, we introduce the adaptive SIF (adaSIF), which takes into account also the used augmentations in its calculation.

\section{Related work}
\label{related_work}
{\bf Membership inference.} Shokri et al. were the first to propose an MI attack against ML models \cite{Shokri2017MembershipIA}. Their attack model includes a bundle of "shadow models" which are trained to mimic the classification output vector of a black-box target model $h$, for training (\textit{members}) and test (\textit{non-members}) samples. These shadow models are then used to generate a shadow dataset. For a given shadow model $S_h^k$ and a sample $(\mathbf{x}_i^k, y_i^k)$, where $\mathbf{x}_i^k$ is an input and $y_i^k$ is its label, they predict the output vector $\mathbf{y}=S_h^k(\mathbf{x})$ and save the record $(y_i^k, \mathbf{y_i^k}, m_i)$, where $m_i$ equals $1$ if $(\mathbf{x}_i^k, y_i^k)$ is a member and $0$ otherwise. The shadow dataset $\Big\{(y_i^k, \mathbf{y_i^k}, m_i)\Big\}_{1\leq i\leq n}^{1\leq k\leq p}$ obtained from $n$ samples and $p$ shadow models is utilized to train a binary classifier as an attack model for the MI prediction.

The aforementioned attack requires training the $S_h^k$ models on similar architecture as $h$, with samples distributed similarly to the training set of $h$. Salem et al. later showed that the exact architecture knowledge is not needed, and any sample distribution of a similar task (e.g., vision task) is sufficient \cite{Salem2019MLLeaksMA}. Moreover, they achieved a comparable MI attack performance using a single shadow model.

Yeom et al. showed that overfitted target models are necessarily vulnerable to MI attacks \cite{Yeom2018PrivacyRI}, and proposed a simple baseline heuristic that predicts a sample $z=(x,y)$ to be a member if the target model prediction $\hat{y}=h(x)$ matches $y$, and a non-member otherwise. We name this baseline the "Gap attack" since its accuracy is correlated with the generalization error, which is the gap between $h$ accuracy on the training data ($A_{mem}$) and the held out data ($A_{non-mem}$):
$$\frac{1}{2} + \frac{1}{2}\big(A_{mem}-A_{non-mem}\big), ~~~\text{where}~ A_{mem},A_{non-mem}\in[0,1].$$

As an attempt to mitigate MI attacks, several defenses were proposed to alter $h$ output confidence vector \cite{Jia2019MemGuardDA,Nasr2018MachineLW}, however recent works presented SOTA MI attack performance on black-box models that only output hard labels, without accessing the class posterior probabilities \cite{ChoquetteChoo2021LabelOnlyMI,DBLP:journals/corr/abs-2007-15528}. To that end, they applied a black-box adversarial attack \cite{Chen2020HopSkipJumpAttackAQ,Li2020QEBAQB} on the input image $x$ image until its label $y$ flipped, and inspected the $L_2$ distance $d=||x-x'||_2$ where $x'$ the adversarial image. Next, they predicted the sample $(x,y)$ to be a member if $d>\tau$ for some threshold $\tau$.

Sablayrolles et al. explored MI attacks in a white-box setting \cite{Sablayrolles2019WhiteboxVB}. They showed that the optimal membership inference only depends on the loss function, and thus claimed that white-box attacks cannot perform better than black-box attacks. Rezaei and Liu also assumed white-box setting and utilized hidden layers activations and gradient norms in their attack models, and observed only a marginal improvement compared to the black-box attack baseline \cite{Rezaei2021OnTD}.

Leino and Fredrikson constructed white-box MI attacks that can be calibrated for its output confidences \cite{leino2020stolen} (the member/non-member classes) and showed that they can obtain higher precision than a black-box attack. However, tuning the MI attack for precision greatly reduced their recall score.
Our work shows that white-box information can assist the adversary and perform SOTA MI, without sacrificing the member recall or the accuracy on the non-member class. \looseness=-1

Nasr et al. utilized a white-box attack that trains a DNN attack model on features collected from all the target model layers, for both the forward pass (activations) and backward pass (gradients) \cite{Nasr2018ComprehensivePA}. Their approach surpassed the performance of a baseline black-box. We show that our attack method achieves even superior results on CIFAR-100 \cite{CIFAR} using their target model training setup.

{\bf Influence functions.}
Koh and Liang proposed to interpret the predictions of an ML model by tracing them through its learning algorithm and training data \cite{Koh17Understanding}. They quantify the influence a train sample $z_{train}$ has on a specific loss value of a test sample $z_{test}$. Aside of interpretability, this measure had been shown to improve classifier training \cite{Shao_Skryagin_Stammer_Schramowski_Kersting_2021}, defend against adversarial attacks \cite{Cohen_2020_CVPR}, and fix mislabeled training data \cite{kong2022resolving}.

The disadvantage of influence functions is that their computation is computationally demanding. To mitigate that, we use the self-influence measure, which calculates the influence an example has on itself and has been used to fix erroneous training labels \cite{NEURIPS2020_e6385d39,Schioppa2021ScalingUI}. It allows us to perform the MI attack in a computationally efficient manner. 

\section{Method}
\label{Method}
In order to describe our approach, we start by formally defining influence functions in general and their derived self-influence functions (SIF) that we use in the paper. Next, we introduce our proposed SIF attack model for neural networks that have been trained without data augmentations. Lastly, we modify our approach to attack target models that are trained with data augmentations.

We study a classification task from an input space $\mathcal{X}$ (e.g., images) to an output space $\mathcal{Y}$ (e.g., labels). For a sample point $z = (x,y) \in \mathcal{X} \times \mathcal{Y}$ and model parameters $\theta$, we denote the loss by $L(z, \theta)$. Let $\big\{z_1,...,z_n\big\}$ be a training set of size $n$, and let $\frac{1}{n}\sum_{i=1}^nL(z_i,\theta)$ be the empirical risk. The empirical risk minimizer is defined by $\hat{\theta} \eqdef \underset{\theta}{\mathrm{argmin}}\frac{1}{n}\sum_{i=1}^nL(z_i,\theta)$. We assume that the empirical risk has first and second gradients and it is strictly convex in $\theta$.

\subsection{Influence functions}
\label{influence_functions}
We study the change in model parameters due to upweighting a specific training sample $z$ by a small $\epsilon$ in the training process. Upweighting $z$ adjusts the model parameters to be $\hat{\theta}_{\epsilon,z} \eqdef \underset{\theta}{\mathrm{argmin}}\frac{1}{n}\sum_{i=1}^nL(z_i,\theta) + \epsilon L(z,\theta)$. 
Cook and Weisberg \cite{Cook1982ResidualsAI} showed that the influence function of upweighting $z$ on the model parameters $\hat{\theta}$ is given by
\begin{equation}
\label{eq:I_up_params}
I_{up,params}(z) \eqdef \frac{d\hat{\theta}_{\epsilon,z}}{d\epsilon}\Big|_{\epsilon=0} = -H_{\hat{\theta}}^{-1}L(z, \theta),
\end{equation}
where $H_{\hat{\theta}}=\frac{1}{n}\sum_{i=1}^n\nabla_{\theta}^2L(z_i,\hat{\theta})$ is the Hessian.

Influence functions interpret an ML model by indicating which of the training samples assisted it to make its prediction, and which training samples were destructive, i.e., inhibited the model from its prediction.
Koh and Liang \cite{Koh17Understanding} proposed to measure the influence a train sample $z_{train}$ has on the loss of a test sample $z$, using the term:
\begin{equation}
\label{eq:I_up_loss}
\begin{split}
I_{up,loss}(z_{train}, z) & \eqdef \frac{dL\big(z,\hat{\theta}_{\epsilon,z_{train}}\big)}{d\epsilon}\Big|_{\epsilon=0} = \nabla_{\theta}L(z,\hat{\theta})^T \frac{d\hat{\theta}_{\epsilon,z_{train}}}{d\epsilon}\Big|_{\epsilon=0} \\
& = -\nabla_{\theta}L(z, \hat{\theta})^TH_{\hat{\theta}}^{-1}\nabla_{\theta}L(z_{train}, \hat{\theta}).
\end{split}
\end{equation}
The influence function $I_{up,loss}(z_{train}, z)$ measures how much the test loss $L(z, \hat{\theta})$ would change if we were to "upweight" the training sample $z_{train}$ in the empirical risk. The influence function is composed of three components: the gradient of the training sample $z_{train}$, the gradient of the test sample $z$, and the "similarity" of these samples with respect to the model perspective that is expressed by the term $H_{\hat{\theta}}^{-1}$, which is a positive definite matrix. In the influence functions formulation, larger gradients and similarity are correlated to larger influence.

\subsection{SIF MI attack}
\label{sec:sif_mi_attack}
Our goal is to build an attack model that is a binary classifier which predicts whether a sample was used to train the target model or not. The hypothesis that underlines our approach is that if an image has been used to train an ML target model, then it would have a large influence measure on test images' loss with the same label. If so, in order to infer whether a specific image is a member (used in training), we need to examine its influence measure (Eq.~\eqref{eq:I_up_loss}) on other images with the same label.

Given an unseen sample $z=(x,y)$ (either member or non-member) and a set of samples known to be non-members $\big\{z_1,\dots,z_m\big\}$ with the same label $y$, a rigorous influence function analysis requires applying Eq.~\eqref{eq:I_up_loss} to every pair $\big(z,z_i\big)$ for $1\leq i \leq m$, and inspect the $m$ obtained influence measures. Alas, the expression $I_{up,loss}(z, z_i)$ requires calculating a Hessian vector product (HVP) and thus it is not scalable for large datasets due to the large computational cost. 
To make our attack model practical with low computational time, we propose a faster approach that utilizes only the influence of a sample $z$ on itself by merely calculating its SIF measure:
\begin{equation}
\label{eq:SIF}
I_{SIF}(z) = -\nabla_{\theta}L(z, \hat{\theta})^TH_{\hat{\theta}}^{-1}\nabla_{\theta}L(z, \hat{\theta}).
\end{equation}
This measure stands for the influence a single sample point has on its own loss.
We calculate $I_{SIF}(z)$ and classify $z$ as member if it satisfies the conditions: (i) $I_{SIF}(z) \in \big[\tau_{min},\tau_{max}\big]$ and (ii) $y = \hat{y}$, where $\hat{y}$ is the prediction of the target model and $\tau_{min},\tau_{max}$ are some thresholds. If any of (i) or (ii) is violated, then we classify $z$ as non-member. A pseudo code of our approach appears in Appendix~\ref{supp_sec:sif_algorithm}.

Notice that our framework operates in the white-box setting, i.e., it requires access to the model's parameters, activations, and to its first/second order gradients. Therefore, it is not a label-only attack and cannot be applied to black-box models.


\subsection{Adaptive attack to augmentations}
\label{sec:adaptive_attack_to_augmentation}
The SIF attack model assumes that a given sample $z$ either belongs to the training set (member) or not (non-member). Alas, most computer vision training schemes employ data augmentation. Thus, the target model might have been introduced to some transformations of the image $x$, instead of the original image. Training with augmentation can be considered as a defense against our SIF based method since Eq.~\eqref{eq:SIF} assumes that a data point $z$ remains unchanged in the training process.
Thus, we propose an adaptation to SIF (Eq.~\eqref{eq:SIF}) to better estimate the influence of a train sample $z$ on itself, assuming that $z$ is augmented during the training.

Calculating the Hessian and its inverse for a DNN is too expensive due to the millions of parameters involved. Note that for $n$ training points and $\theta\in\mathbb{R}^p$, this calculation has a complexity of $\mathrm{O\big(np^2+p^3\big)}$. To overcome this problem, we avoid the explicit calculation of $H_{\theta}^{-1}$ and use HVPs with stochastic estimation, as proposed by \cite{Koh17Understanding}.
Specifically, we approximate the vector $\mathrm{s}(z) = H_{\theta}^{-1}\nabla_{\theta}L(z, \theta)$ using a stochastic estimation method proposed by \cite{Agarwal2016SecondOS}
and then rewrite Eq.~\eqref{eq:SIF} as: $$I_{SIF}(z) = -\mathrm{s}(z) \cdot \nabla_{\theta}L(z, \theta).$$

With this approximation at hand, we turn to describe our adaptive attack to augmentations, adaSIF.
Let $z=\big(x, y\big)$ denote an original sample and $I$ be a random data augmentation operator sampled from the family of training augmentation distribution $\mathcal{T}$ $\big(I\sim\mathcal{T}\big)$. Then, we define the adaptive self-influence measure of $z$ on Eq.~\eqref{eq:SIF} as:
\begin{equation}
\label{eq:adaSIF}
I_{adaSIF}(z) = -\mathrm{s}(z) \cdot \mathbb{E}_{I\sim\mathcal{T}}\Big[\nabla_{\theta}L\big(I(x), y, \theta\big)\Big].
\end{equation}
Note that in adaSIF, we average the influence of different augmentations of $z$ on itself. For calculating the term $\mathbb{E}_{I\sim\mathcal{T}}\Big[\nabla_{\theta}L\big(I(x), y, \theta\big)\Big]$, we followed the same implementation as used in \cite{Koh17Understanding}, but instead of sampling training set samples (as the goal in \cite{Koh17Understanding} was to the check influence of the training examples on $z$), we sampled different augmentations of $z$, $I(z)$, as our goal is to check the influence of the augmentations on $z$.
We compared adaSIF with a naive ensemble of SIF measures calculated on data augmentations assemble and found that adaSIF is slightly better in most cases. For more details on adaSIF see Appendix~\ref{supp_sec:efficient_sif_calculation}. A comparison between adaSIF to the naive ensemble is shown in Appendix~\ref{supp_sec:naive_sif_ensemble}.


\section{Experimental setup}
\label{sec:experimental_setup}
Here we list the seven target models we used for evaluating our work, and provide technical details on how they were trained. We then describe the dataset split done to fit our attack model and present the balanced accuracy metric used to compare between all attack models. The CPU and GPU apparatus used in our experiments is described in Appendix~\ref{supp_sec:hardware_setup}.

\subsection{Target model and implementation details}
\label{sec:target_model}
Since overfitted machine learning models are more susceptible to membership leakage \cite{Shokri2017MembershipIA,Salem2019MLLeaksMA,Song2019PrivacyRO}, we trained seven different target models $\mathcal{M}$-1, ..., $\mathcal{M}$-7, where each model differs only by the training set size. A similar target model setup was also utilized in previous works \cite{DBLP:journals/corr/abs-2007-15528,Truex2018TowardsDM}. The sizes of the target models are summarized in Table~\ref{tab:target_models}. The Tiny ImageNet \cite{TinyImageNet} dataset was not evaluated on $\mathcal{M}$-1 since it has 200 labels which exceed $\mathcal{M}$-1 training set size.
\begin{table}
\caption{The number of the training set size \big|$\mathcal{D}_{mem}$\big| for each of the target models.}
\centering
\begin{tabular}{c|ccccccc}
\toprule
Target Model & $\mathcal{M}$-1 & $\mathcal{M}$-2 & $\mathcal{M}$-3 & $\mathcal{M}$-4 & $\mathcal{M}$-5 & $\mathcal{M}$-6 & $\mathcal{M}$-7 \\ \big|$\mathcal{D}_{mem}$\big| & 100 & 1000 & 5000 & 10000 & 15000 & 20000 & 25000 \\
\bottomrule
\end{tabular}
\label{tab:target_models}
\vspace{-0.18in}
\end{table}

We split the full official training set of CIFAR-10, CIFAR-100, and Tiny ImageNet into \textit{training} and \textit{validation} sets. The \textit{training} size is set by Table~\ref{tab:target_models} and \textit{validation} was set to $5\%$ of the official training set. 
Three DNN architectures were used in our experiments to train the target models: Resnet18 \cite{RESNET}, AlexNet \cite{Krizhevsky2012ImageNetNetworks}, and DenseNet \cite{DenseNet}. 
We applied ReLU activations for all models and optimized the cross entropy loss while decaying the learning rate using the \textit{validation} set's accuracy score, for 400 epochs, batch size 100, with $L_2$ weight regularization of 0.0001, using a stochastic gradient decent optimizer with momentum 0.9 and Nesterov updates. We did not include batch-norm weights into the $L_2$ weight regularization. For the data augmentation adaptive attack in Section~\ref{sec:data_augmentation_adaptive_attack} we trained the target models with random crop and horizontal flipping. We used the model checkpoint with the best (highest) accuracy on the validation set. The full DNN \textit{training}, \textit{validation}, and official test accuracies of the target models are reported in Appendix~\ref{supp_sec:accuracy_of_target_models}.

\subsection{Attack model training and evaluation}
\label{sec:attack_model_training_and_evaluation}
To train and evaluate our SIF attack model, we split each dataset into $\mathcal{D}_{mem}$ and $\mathcal{D}_{non-mem}$ subsets. The former is the \textit{training} set defined in Section~\ref{sec:target_model}, whereas the latter holds only images that are outside the \textit{training} and \textit{validation} sets.
$\mathcal{D}_{mem}$ and $\mathcal{D}_{non-mem}$ were further divided to $\mathcal{D}_{mem}^{train}$, $\mathcal{D}_{non-mem}^{train}$, and $\mathcal{D}_{mem}^{test}$, $\mathcal{D}_{non-mem}^{test}$, where the first two subsets were used to fit the attack models and the last two subsets were used to evaluate membership inference by the attack models.
For simplicity, we matched the test set size to the training set size, more explicitly we set \big|$\mathcal{D}_{mem}^{train}$\big| = \big|$\mathcal{D}_{mem}^{test}$\big| = \big|$\mathcal{D}_{non-mem}^{train}$\big| = \big|$\mathcal{D}_{non-mem}^{test}$\big|.

The attack model's thresholds $\tau_{min},\tau_{max}$ (Section~\ref{sec:sif_mi_attack}) are chosen to optimize the Balanced accuracy in Eq.~\eqref{eq:balanced_accuracy} on $\mathcal{D}_{mem}^{train}$ and $\mathcal{D}_{non-mem}^{train}$, similarly to \cite{DBLP:journals/corr/abs-2007-15528}. The threshold choosing algorithm is provided in Algorithm~\ref{alg:set_thresholds} in Appendix~\ref{supp_sec:sif_algorithm}. Since the SIF attack requires setting two thresholds, we choose to evaluate our MI attack using the balanced accuracy as done in \cite{Rezaei2021OnTD,ChoquetteChoo2021LabelOnlyMI} instead of the AUC of the ROC curve. 
We denote $N_1=\big|\mathcal{D}_{mem}^{test}\big|$, and $N_2=\big|\mathcal{D}_{non-mem}^{test}\big|$. Our MI test samples are denoted as $\mathcal{D}_{mem}^{test}=\Big\{(x_m^1,y_m^1), ..., (x_m^{N_1},y_m^{N_1})\Big\}$ and $\mathcal{D}_{non-mem}^{test}=\Big\{(x_{nm}^1,y_{nm}^1), ..., (x_{nm}^{N_2},y_{nm}^{N_2})\Big\}$, where $x$ denotes an image and $y_m,y_{nm}$ labels denote member (1), non-member (0) labels, respectively. 

The balanced accuracy of an attack model is then defined by:
\begin{equation}
\label{eq:balanced_accuracy}
\text{Balanced Acc} = \frac{1}{N_1+N_2}\Bigg[\sum_{i=1}^{N1}\hat{y}_m^i + \sum_{i=1}^{N2}\big(1-\hat{y}_{nm}^i\big)\Bigg],
\end{equation}
where $\hat{y}_m^i$ and $\hat{y}_{nm}^i$ are the attack model's predictions for $y_m^i$ and $y_{nm}^i$, respectively. 

For the baseline comparison, in our experiments we used the Gap, Black-box, and Boundary distance MI attacks implementation from ART\footnote{\href{https://github.com/Trusted-AI/adversarial-robustness-toolbox}{https://github.com/Trusted-AI/adversarial-robustness-toolbox}} \cite{Nicolae2018AdversarialRT}. The Boundary distance attack was implemented with the HopSkipJump adversarial attack \cite{Chen2020HopSkipJumpAttackAQ}. Due to a very long computation time of the Boundary distance attack and our adaSIF attack (Section~\ref{sec:adaptive_attack_to_augmentation}), we limited the size of the fitting and evaluation subsets to 1000 and 5000, respectively, for these attack models only.

\section{Results}
\label{Results}
We start by presenting histograms for SIF and adaSIF values to motivate the use of our method. We then evaluate the performance of our SIF MI attack and compare it to current SOTA attack methods. Next, we conduct ablation studies aimed to improve the adaSIF attack with minimal run time. Next, we test our adaSIF attack on target models trained with data augmentations, a defense that aims to mitigate our vanilla SIF attack.
Lastly, we show the fitting and inference time for all the attack models used in this paper.

\subsection{SIF distribution of membership}
\begin{figure}[ht]
\centering
\includegraphics[width=\linewidth]{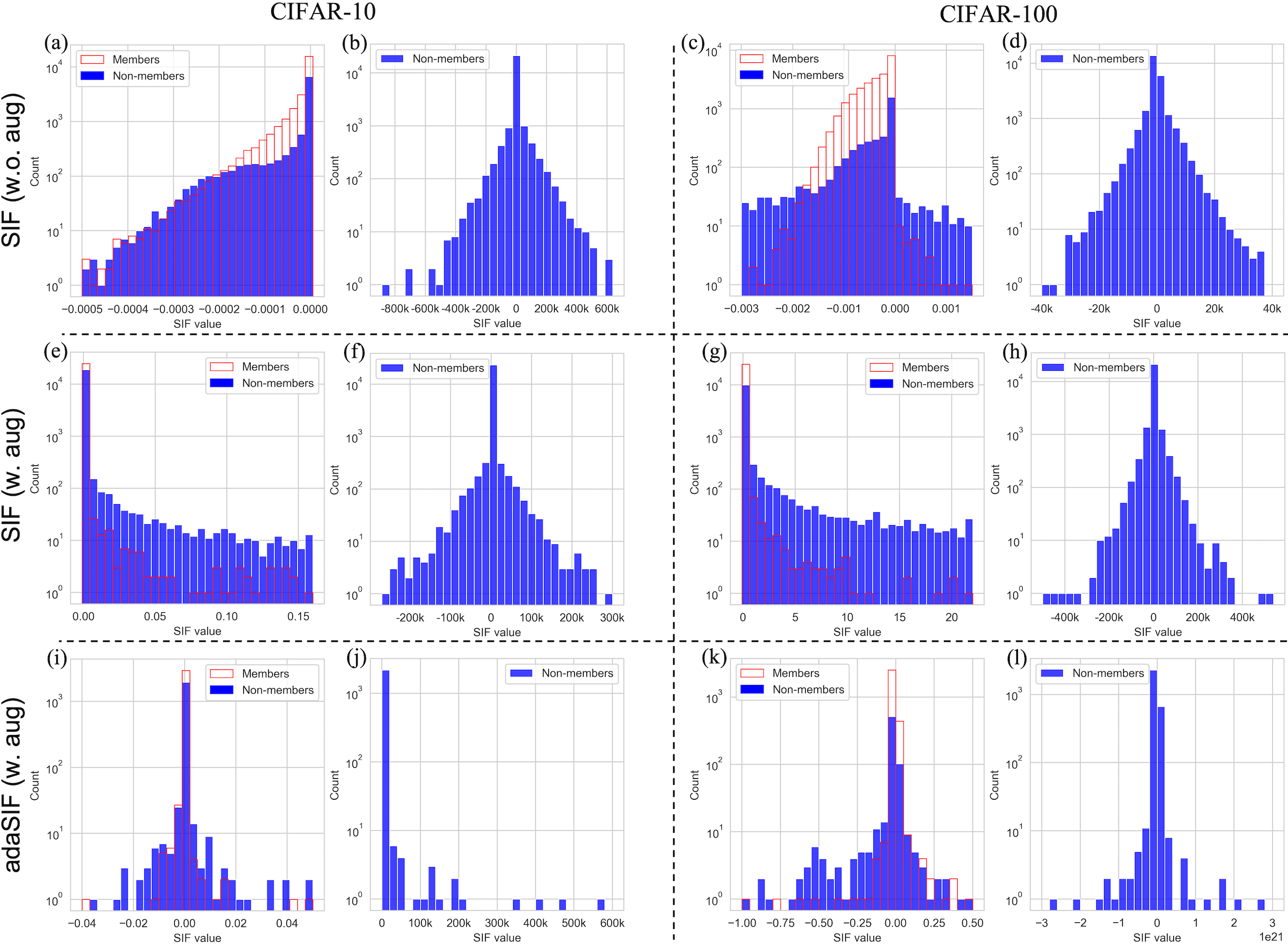}
\vspace{-0.2in}
\caption{SIF and adaSIF values distribution for CIFAR-10 and CIFAR-100 images within the training set (\textit{members}) and images outside it (\textit{non-members}) for target model $\mathcal{M}$-7 trained on Resnet18.
The top and middle rows correspond to SIF values (Eq.~\eqref{eq:SIF}) obtained from models trained without and with data augmentation, respectively. The bottom row corresponds to adaSIF values (Eq.~\eqref{eq:adaSIF}) obtained from a model trained with data augmentations. All members reside in a short range near $0$; non-members seldom share the same distribution in this range, and they can attain extreme values. This information is exploited by our attack. (e) and (g) show that data augmentations expands the SIF values for members and mitigate our vanilla SIF attack. (i) and (k) show that applying adaSIF can recover the short range property for the members and therefore facilitate the attack.}
\label{fig:sif_hist}
\vspace{-0.1in}
\end{figure}

To better understand how our attack works, we show in Figure~\ref{fig:sif_hist} the SIF values (Eq.~\eqref{eq:SIF}) distribution for members and non-members of CIFAR-10 and CIFAR-100, calculated on the target model $\mathcal{M}$-7 trained on Resnet18. The top row shows SIF values on a model trained without data augmentations. We observe that members are distributed solely within a short interval around $0$, whereas non-members can attain very large absolute values, and their distribution on the aforementioned interval seldom matches the members' distribution. This shows that member samples have negligible influence scores on themselves, while non-member samples have a large impact on their test loss. Our SIF attack exploit that property and sets thresholds $\tau_1$ and $\tau_2$ to encapsulate most of the members.

The middle row shows the same SIF values when calculated on a model trained with data augmentations. In this case the non-members still exhibit extreme values, but the members' range spans to a larger interval (see Figure~\ref{fig:sif_hist}(e) and Figure~\ref{fig:sif_hist}(g) compared to Figure~\ref{fig:sif_hist}(a) and Figure~\ref{fig:sif_hist}(c), respectively). Thus, data augmentation can be considered as a defense to SIF since it requires setting an expanded range [$\tau_2,\tau_1$] which hampers our attack.

The bottom row shows adaSIF values (Eq~\eqref{eq:adaSIF}) when calculated on the same data augmented target model that is used in the middle row. We observe that adaSIF restores the short range characteristic for the members, and therefore negates the effect of the data augmentation on the target model defense.

\subsection{Comparison of MI attacks}
\label{sec:comparison_of_mi_attacks}
\begin{figure}[ht!]
\includegraphics[width=1\linewidth]{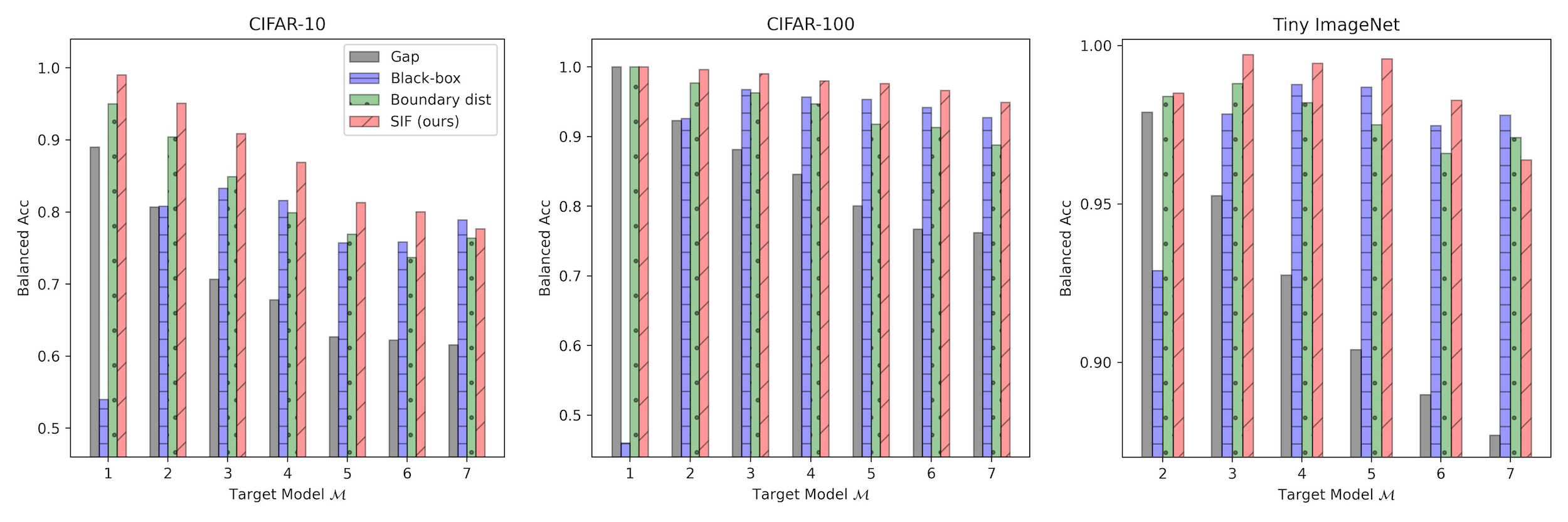}
\vspace{-0.24in}
\caption{Comparison of our SIF attack with some baseline MI attacks: Gap, Black-box, and Boundary distance. The x-axis indicates the attacked target model and the y-axis shows the balanced attack accuracy (Eq.~\eqref{eq:balanced_accuracy}). Our SIF method surpasses previous SOTA attacks for most target models.}
\label{fig:all_attack_scores_small}
\vspace{-0.18in}
\end{figure}

Figure~\ref{fig:all_attack_scores_small} shows the attack power (balanced accuracy) of the four inspected attacks: Gap (black), Black-box (blue), Boundary distance (green), and SIF (red), on three popular classification tasks: CIFAR-10, CIFAR-100, and Tiny ImageNet. We compare between the attack scores calculated on seven different Resnet18 target models (Table~\ref{tab:target_models}), where each model was trained on different number of samples. Our SIF attack achieves higher MI accuracy than the baselines for most of the target models. Table~\ref{tab/all_attack_scores} summarizes the attack scores for all the MI methods presented in Figure~\ref{fig:all_attack_scores_small}, and also details both the member and non-member accuracies. We observe that SIF almost always achieves perfect accuracy ($\sim1.0$) for the members, which is crucial for a reliable membership inference. We run the same comparison for AlexNet and DenseNet in Appendix~\ref{supp_sec:comparison_of_mi_attacks} and show that SIF achieves new SOTA for these architectures as well. A more detailed analysis with precision and recall values is presented in Apprndix~\ref{supp_sec:precision_and_recall}.

\begin{table*}[ht!]
\caption{Comparison of accuracies for various MI attack methods: Gap, Black-box, Boundary distance, and SIF. We detail for every attack the accuracy on the members, the non-members, and the balanced accuracy.}
\resizebox{1\columnwidth}{!}{
\begin{tabular}{cc|ccc|ccc|ccc|ccc}
\toprule
\multirow{2}{*}{Dataset} & \multirow{2}{*}{\parbox{1cm}{Target model}} & \multicolumn{3}{c}{Gap} & \multicolumn{3}{c}{Black-box} & \multicolumn{3}{c}{Boundary dist} & \multicolumn{3}{c}{SIF (ours)} \\
& & Member & Non-mem & Balanced & Member & Non-mem & Balanced & Member & Non-mem & Balanced & Member & Non-mem & Balanced\\
\hline
\multirow{7}{*}{CIFAR-10} & $\mathcal{M}$-1 & 1.000 & 0.780 & 0.890 & 0.600 & 0.480 & 0.540 & 0.980 & 0.920 & 0.950 & 1.000 & 0.980 & \textbf{0.990} \\    
& $\mathcal{M}$-2 & 1.000 & 0.614 & 0.807 & 1.000 & 0.616 & 0.808 & 0.994 & 0.814 & 0.904 & 0.996 & 0.906 & \textbf{0.951} \\
& $\mathcal{M}$-3 & 1.000 & 0.414 & 0.707 & 1.000 & 0.666 & 0.833 & 0.946 & 0.752 & 0.849 & 1.000 & 0.818 & \textbf{0.909} \\
& $\mathcal{M}$-4 & 1.000 & 0.356 & 0.678 & 0.986 & 0.646 & 0.816 & 0.914 & 0.684 & 0.799 & 0.989 & 0.749 & \textbf{0.869} \\
& $\mathcal{M}$-5 & 1.000 & 0.254 & 0.627 & 1.000 & 0.515 & 0.757 & 0.950 & 0.588 & 0.769 & 0.987 & 0.639 & \textbf{0.813} \\
& $\mathcal{M}$-6 & 1.000 & 0.244 & 0.622 & 0.886 & 0.631 & 0.758 & 0.970 & 0.504 & 0.737 & 0.976 & 0.624 & \textbf{0.800} \\
& $\mathcal{M}$-7 & 1.000 & 0.231 & 0.616 & 1.000 & 0.578 & \textbf{0.789} & 0.910 & 0.618 & 0.764 & 1.000 & 0.553 & 0.777 \\
\hline
\multirow{7}{*}{CIFAR-100} & $\mathcal{M}$-1 & 1.000 & 1.000 & \textbf{1.000} & 0.140 & 0.780 & 0.460 & 1.000 & 1.000 & \textbf{1.000} & 1.000 & 1.000 & \textbf{1.000} \\
& $\mathcal{M}$-2 & 1.000 & 0.846 & 0.923 & 1.000 & 0.852 & 0.926 & 1.000 & 0.954 & 0.977 & 0.998 & 0.994 & \textbf{0.996} \\
& $\mathcal{M}$-3 & 1.000 & 0.763 & 0.882 & 1.000 & 0.935 & 0.967 & 0.988 & 0.938 & 0.963 & 0.999 & 0.982 & \textbf{0.990} \\
& $\mathcal{M}$-4 & 1.000 & 0.692 & 0.846 & 1.000 & 0.913 & 0.957 & 0.998 & 0.896 & 0.947 & 1.000 & 0.960 & \textbf{0.980} \\
& $\mathcal{M}$-5 & 1.000 & 0.601 & 0.801 & 1.000 &	0.907 & 0.953 & 0.974 & 0.862 & 0.918 & 1.000 & 0.953 & \textbf{0.976} \\
& $\mathcal{M}$-6 & 1.000 & 0.535 & 0.767 & 0.993 & 0.891 & 0.942 & 0.986 & 0.840 & 0.913 & 1.000 & 0.932 & \textbf{0.966} \\
& $\mathcal{M}$-7 & 1.000 & 0.524 & 0.762 & 0.993 & 0.861 & 0.927 & 0.978 & 0.798 & 0.888 & 0.999 & 0.900 & \textbf{0.949} \\
\hline
\multirow{6}{*}{Tiny ImageNet} & $\mathcal{M}$-2 & 0.996 & 0.962 & 0.979 & 0.944 & 0.914 & 0.929 & 0.992 & 0.976 & 0.984 & 0.992 & 0.978 & \textbf{0.985} \\
& $\mathcal{M}$-3 & 1.000 & 0.905 & 0.953 & 1.000 & 0.957 & 0.978 & 1.000 & 0.976 & 0.988 & 1.000 & 0.994 & \textbf{0.997} \\
& $\mathcal{M}$-4 & 1.000 & 0.855 & 0.928 & 1.000 & 0.976 & 0.988 & 1.000 & 0.964 & 0.982 & 1.000 & 0.989 & \textbf{0.994} \\
& $\mathcal{M}$-5 & 1.000 & 0.808 & 0.904 & 1.000 & 0.974 & 0.987 & 0.992 & 0.958 & 0.975 & 1.000 & 0.992 & \textbf{0.996} \\
& $\mathcal{M}$-6 & 1.000 & 0.780 & 0.890 & 0.999 &	0.950 & 0.975 & 0.988 & 0.944 & 0.966 & 1.000 & 0.966 & \textbf{0.983} \\
& $\mathcal{M}$-7 & 1.000 & 0.754 & 0.877 & 0.994 & 0.962 & \textbf{0.978} & 0.996 & 0.946 & 0.971 & 1.000 & 0.928 & 0.964 \\
\bottomrule
\end{tabular}
}
\label{tab/all_attack_scores}
\vspace{-0.1in}
\end{table*}

\subsection{Adaptive attack ablation study}
To better understand the impact of the different terms on our method, we perform several ablation experiments. 
The adaSIF attack described in Section~\ref{sec:adaptive_attack_to_augmentation} requires a proper approximation of the $s(z)$ term in Eq.~\eqref{eq:adaSIF}; this approximation is controlled by two parameters: (i) $r$, the number of iterations used to estimate $s(z)$; and (ii) the recursion depth $d$, i.e., the number of augmentations to perform during one iteration of $s(z)$ calculation. Increasing either parameter prolongs the attack's inference time so we aim for the smallest values of $r$, $d$ for a successful adaptive attack.

\begin{figure}[ht]
\centering
\includegraphics[width=0.8\linewidth]{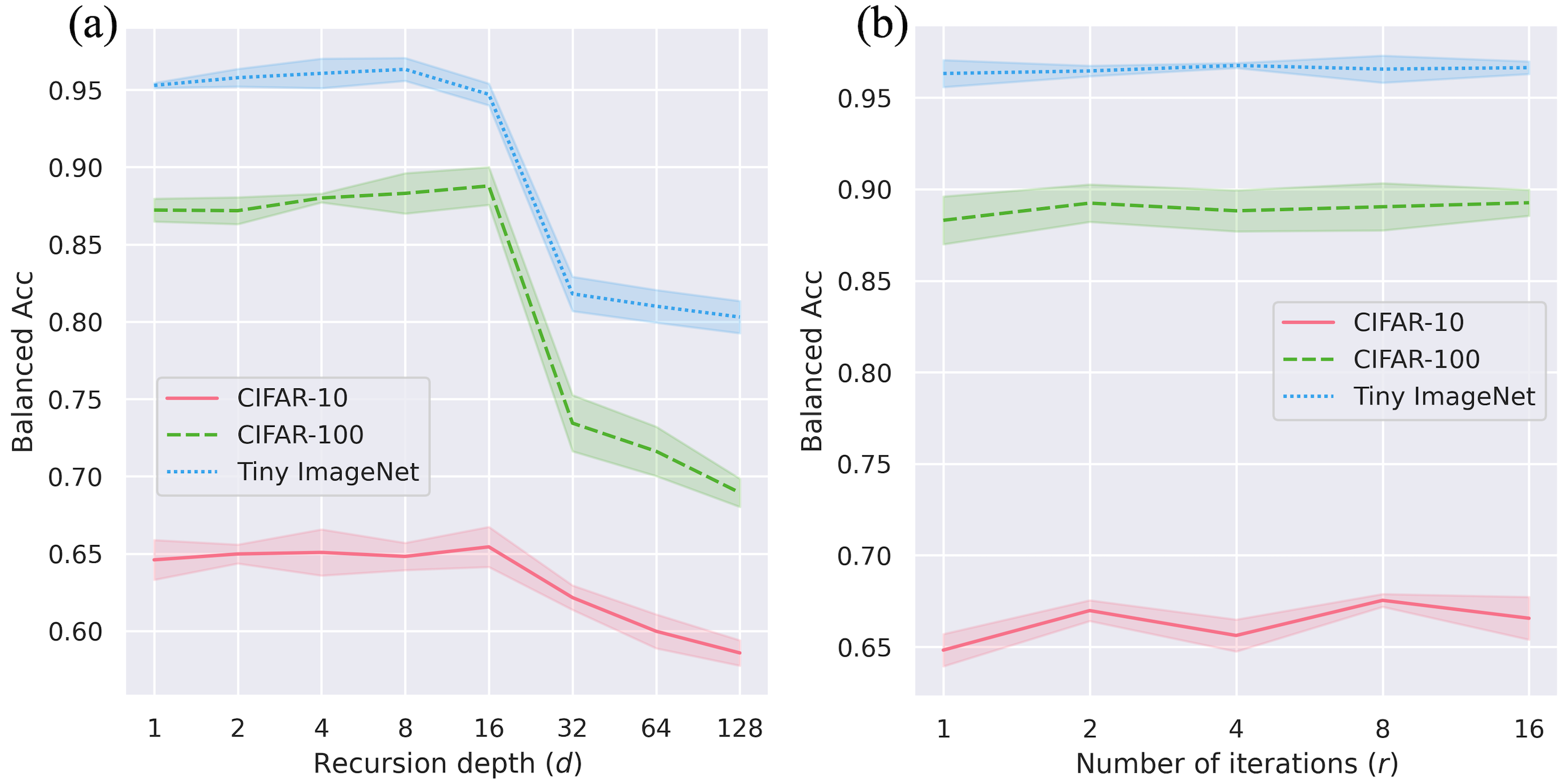}
\vspace{-0.1in}
\caption{Ablation study on the recursion depth ($d$) and number of iterations ($r)$ used to estimate $s(z)$ in Eq.~\eqref{eq:adaSIF}. The balanced accuracy of our adaSIF attack was calculated for target model $\mathcal{M}$-7 as a function of: (a) $d$ where $r=1$ and (b) $r$ where $d=8$. Both $d$ and $r$ are shown in logarithmic scale.}
\label{fig:ablation}
\vspace{-0.18in}
\end{figure}

Figure~\ref{fig:ablation}(a) shows the effect of $d$ on the balanced accuracy of our adaptive adaSIF attack, for CIFAR-10, CIFAR-100, and Tiny ImageNet trained on target model $\mathcal{M}$-7 with $r$ set to $1$. The width of each line corresponds to the measured standard deviation of five experiments. We set adaSIF with $d=8$ since it achieves a good balanced accuracy with high confidence (narrow interval).

Next, we inspect the effect of $r$ on the balanced accuracy with $d$ set to $8$. Figure~\ref{fig:ablation}(b) shows that $r$ has a marginal impact on the balanced accuracy for CIFAR-100 and Tiny ImageNet and some improvement for CIFAR-10. We therefore select $d=8$ and $r=8$ for our adaSIF method. Yet, one may gain very similar results using our attack by using $r=1$, which reduces the computational time by a factor of $8$. 

\subsection{Data augmentation adaptive attack}
\label{sec:data_augmentation_adaptive_attack}
\begin{figure}[ht]
\includegraphics[width=1\linewidth]{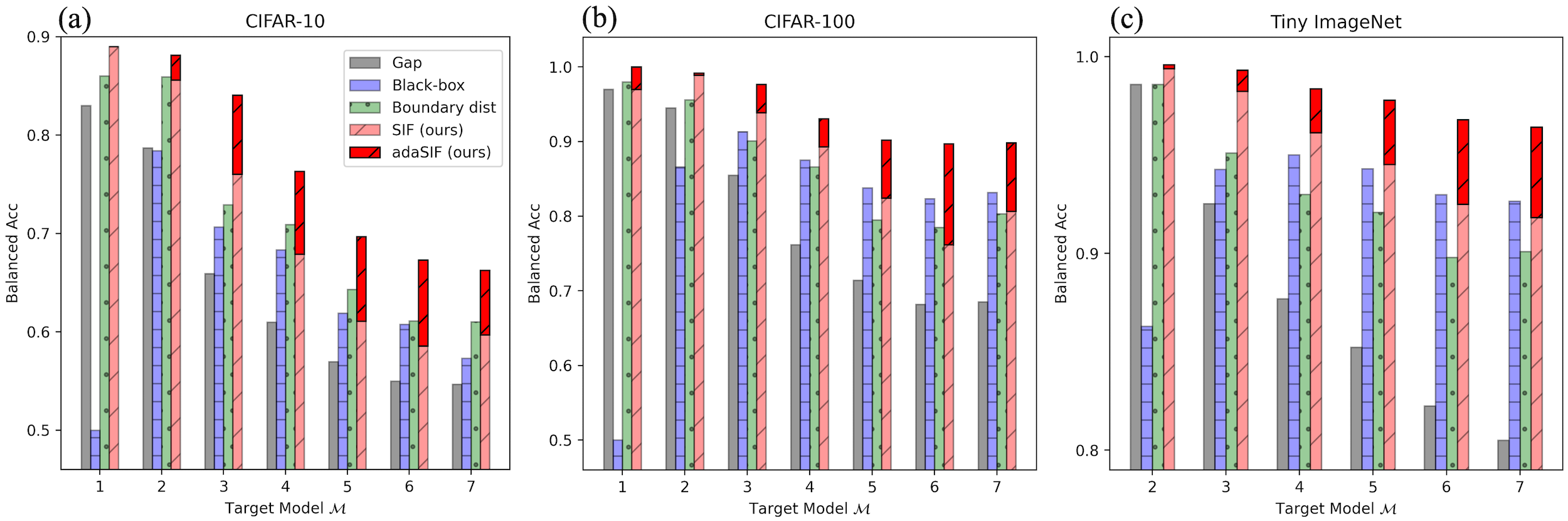}
\vspace{-0.2in}
\caption{Similar comparison of MI attack as in Figure~\ref{fig:all_attack_scores_small} when the target models are trained with data augmentation. The vanilla SIF value from Eq.~\eqref{eq:SIF} performs on par to previous SOTA. When implementing the adaptive attack from Eq.~\eqref{eq:adaSIF} (red bar) we surpass all previous attacks by a large margin.   \looseness=-1}
\label{fig:adaptive_attack_scores_small}
\vspace{-0.18in}
\end{figure}

We repeat the same comparison of MI attacks in Section~\ref{sec:comparison_of_mi_attacks}, where the target models are trained with data augmentation.  Figure~\ref{fig:adaptive_attack_scores_small} shows the balanced accuracy (Eq.~\ref{eq:balanced_accuracy}) of the attacks: Gap, Black-box, Boundary distance, SIF, and our adaptive adaSIF, on CIFAR-10, CIFAR-100, and Tiny ImageNet, for the different target models trained on Resnet18. As expected, our vanilla SIF attack efficacy is attenuated and surpassed by a baseline in most cases. On the other hand, adaSIF boosts our SIF attack to a new SOTA (red bar) for all datasets. Similar results for AlexNet and DenseNet are shown in Appendix~\ref{supp_sec:comparison_of_mi_attacks_with_data_augmentation}.

In another experiment, we trained target models for CIFAR-100 with the full training set of 50000 samples, similarly to \cite{Nasr2018ComprehensivePA}. We show that our adaSIF attack surpasses their reported white-box MI attack accuracy (see Appendix~\ref{supp_sec:comparison_to_a_white_box_attack}).

\subsection{Computational cost}
\label{sec:computational_cost}
Computation time is particularly an issue for calculating the HVP values for the influence function in large datasets \cite{Koh17Understanding}. Table~\ref{tab:computational_cost} presents a comparison of the fitting and inference time for our SIF attack, adaptive adaSIF attack (with $d=8$, $r=8$), and the other baseline attacks used in our experiments, on Tiny ImageNet trained using the $\mathcal{M}$-7 target model. "Fitting" indicates the time used to fine-tune the attack model's parameters, and "inference" indicates the average cost time for a membership inference on a single data point.

The Gap attack model has no parameters and thus does not need fitting. In addition, Gap and Black-box attacks have negligible inference time. Since the Boundary distance and adaSIF attacks are very slow, we fitted and evaluated them on 1000 random samples from $\mathcal{D}_{mem}^{train} \cup \mathcal{D}_{non-mem}^{train}$ and on 5000 random samples from $\mathcal{D}_{mem}^{test} \cup \mathcal{D}_{non-mem}^{test}$, respectively. The vanilla SIF was fitted and evaluated on all the attack model's samples, similarly to Gap and Black-box. We observe that SIF and adaSIF take less time to fit than the Boundary distance, and their inference cost is also lower, particularly for the SIF attack which runs a single MI attack in less than a second.

\begin{table*}[ht!]
\caption{Time required for fine-tuning ("fitting") an attack model on its training points and running a single membership inference ("inference"). Our SIF attack takes a considerable amount of time to fit, but its inference time is much shorter than the Boundary distance attack.}
\resizebox{1\columnwidth}{!}{
\begin{tabular}{c|cc|cc|cc|cc|cc}
\toprule
\multirow{2}{*}{Architecture} & \multicolumn{2}{c}{Gap} & \multicolumn{2}{c}{Black-box} & \multicolumn{2}{c}{Boundary dist} & \multicolumn{2}{c}{SIF (ours)} & \multicolumn{2}{c}{adaSIF (ours)} \\
& Fitting & Inference & Fitting & Inference & Fitting & Inference & Fitting & Inference & Fitting & Inference \\
\hline
Resnet18 & - & 0.19 ms & 165.28 s & 0.22 ms & 7.87 hr & 22.51 s & 4.67 hr & 0.66 s & 5.31 hr & 16.11 s \\
AlexNet  & - & 0.10 ms & 169.35 s & 0.14 ms & 6.68 hr & 20.40 s & 4.19 hr & 0.62 s & 4.27 hr & 12.20 s \\
DenseNet & - & 0.13 ms & 169.26 s & 0.20 ms & 8.41 hr & 25.57 s & 4.72 hr & 0.68 s & 7.42 hr & 16.39 s \\
\bottomrule
\end{tabular}
}
\label{tab:computational_cost}
\end{table*}

{\bf Limitations.} Our attack methods are limited to the white-box threat model only, and an adversary cannot apply them on black-box environments (e.g., ML as a service \cite{7424435}). In addition, our adaSIF attack cannot operate in real-time since its inference time on a single data point exceeds $10$ seconds.

{\bf Negative social impact.} Our work provides a tool for adversaries to steal information from target models. More explicitly, we grant attackers the knowledge if a data point is a member of the training set or not. This unauthorized use of sensitive data breaches user privacy and violates the GDPR regulation. \looseness=-1
\section{Conclusions}
\label{sec:conclusions}
In this paper we addressed the task of membership inference, which is the prediction whether a data sample was used to train a model or not. We showed that the self-influence function in Eq.~\eqref{eq:SIF} is an excellent indicator for membership inference. The aforementioned SIF values combined with the target model's prediction were used to achieve new SOTA MI attack performance for CIFAR-10, CIFAR-100, and Tiny ImageNet, on Resnet18, AlexNet, and Densenet, for various target models (Table~\ref{tab:target_models}).    \looseness=-1

Furthermore, we showed that our SIF attack can be adjusted to address the common MI defense of training the target model with data augmentation. This refined adaSIF attack surpasses all other baselines by a large margin, for every dataset, architecture and target model listed above, while requiring an inference time of $15$ seconds. 

One possible direction to form a more sophisticated defense method against our SIF and adaSIF attacks is to "shift" the members distributions towards the non-members. Yet, this involves Hessian estimation which makes such a method computationally demanding.


{
\small
\bibliographystyle{plain}
\bibliography{my_bib}

\begin{thebibliography}{10}

\bibitem{Agarwal2016SecondOS}
Naman Agarwal, Brian Bullins, and Elad Hazan.
\newblock {Second Order Stochastic Optimization in Linear Time}.
\newblock {\em ArXiv}, abs/1602.0, 2016.

\bibitem{CALLI2021102125}
Erdi {\c{C}}allı, Ecem Sogancioglu, Bram van Ginneken, Kicky~G van Leeuwen,
  and Keelin Murphy.
\newblock {Deep learning for chest X-ray analysis: A survey}.
\newblock {\em Medical Image Analysis}, 72:102125, 2021.

\bibitem{Chen2020HopSkipJumpAttackAQ}
Jianbo Chen and Michael~I Jordan.
\newblock {HopSkipJumpAttack: A Query-Efficient Decision-Based Attack}.
\newblock {\em 2020 IEEE Symposium on Security and Privacy (SP)}, pages
  1277--1294, 2020.

\bibitem{ChoquetteChoo2021LabelOnlyMI}
Christopher~A Choquette-Choo, Florian Tram{\`{e}}r, Nicholas Carlini, and
  Nicolas Papernot.
\newblock {Label-Only Membership Inference Attacks}.
\newblock In {\em ICML}, 2021.

\bibitem{Cohen_2020_CVPR}
Gilad Cohen, Guillermo Sapiro, and Raja Giryes.
\newblock Detecting adversarial samples using influence functions and nearest
  neighbors.
\newblock In {\em CVPR}, 2020.

\bibitem{Cook1982ResidualsAI}
R~Dennis Cook and Sanford Weisberg.
\newblock {Residuals and Influence in Regression}.
\newblock 1982.

\bibitem{Damrongsakmethee2017DataMA}
Thitimanan Damrongsakmethee and Victor-Emil Neagoe.
\newblock {Data Mining and Machine Learning for Financial Analysis}.
\newblock {\em Indian journal of science and technology}, 10:1--7, 2017.

\bibitem{Devlin2019BERTPO}
Jacob Devlin, Ming-Wei Chang, Kenton Lee, and Kristina Toutanova.
\newblock {BERT: Pre-training of Deep Bidirectional Transformers for Language
  Understanding}.
\newblock {\em ArXiv}, abs/1810.0, 2019.

\bibitem{RESNET}
Kaiming He, Xiangyu Zhang, Shaoqing Ren, and Jian Sun.
\newblock {Deep Residual Learning for Image Recognition}.
\newblock {\em CVPR}, pages 770--778, 2016.

\bibitem{DenseNet}
Gao Huang, Zhuang Liu, and Kilian~Q Weinberger.
\newblock {Densely Connected Convolutional Networks}.
\newblock {\em CVPR}, pages 2261--2269, 2017.

\bibitem{Jia2019MemGuardDA}
Jinyuan Jia, Ahmed Salem, Michael Backes, Yang Zhang, and Neil~Zhenqiang Gong.
\newblock {MemGuard: Defending against Black-Box Membership Inference Attacks
  via Adversarial Examples}.
\newblock {\em Proceedings of the 2019 ACM SIGSAC Conference on Computer and
  Communications Security}, 2019.

\bibitem{Koh17Understanding}
Pang~Wei Koh and Percy Liang.
\newblock {Understanding Black-box Predictions via Influence Functions}.
\newblock In {\em ICML}, volume~70, pages 1885--1894, 2017.

\bibitem{kong2022resolving}
Shuming Kong, Yanyan Shen, and Linpeng Huang.
\newblock {Resolving Training Biases via Influence-based Data Relabeling}.
\newblock In {\em International Conference on Learning Representations}, 2022.

\bibitem{CIFAR}
Alex Krizhevsky.
\newblock {Learning multiple layers of features from tiny images}.
\newblock Technical report, 2009.

\bibitem{Krizhevsky2012ImageNetNetworks}
Alex Krizhevsky, Ilya Sutskever, and Geoffrey~E Hinton.
\newblock {ImageNet Classification with Deep Convolutional Neural Networks}.
\newblock {\em NeurIPS}, pages 1--9, 2012.

\bibitem{TinyImageNet}
Ya~Le and X~Yang.
\newblock {Tiny ImageNet Visual Recognition Challenge}.
\newblock 2015.

\bibitem{leino2020stolen}
Klas Leino and Matt Fredrikson.
\newblock {Stolen Memories: Leveraging Model Memorization for Calibrated
  {\$}{\textbackslash}{\{}{\$}White-Box{\$}{\textbackslash}{\}}{\$} Membership
  Inference}.
\newblock In {\em 29th USENIX Security Symposium (USENIX Security 20)}, pages
  1605--1622, 2020.

\bibitem{Li2020QEBAQB}
Huichen Li, Xiaojun Xu, Xiaolu Zhang, Shuang Yang, and Bo~Li.
\newblock {QEBA: Query-Efficient Boundary-Based Blackbox Attack}.
\newblock {\em CVPR}, pages 1218--1227, 2020.

\bibitem{Li2020MembershipIA}
Jiacheng Li, Ninghui Li, and Bruno Ribeiro.
\newblock {Membership Inference Attacks and Defenses in Supervised Learning via
  Generalization Gap}.
\newblock {\em ArXiv}, abs/2002.1, 2020.

\bibitem{DBLP:journals/corr/abs-2007-15528}
Zheng Li and Yang Zhang.
\newblock {Label-Leaks: Membership Inference Attack with Label}.
\newblock {\em CoRR}, abs/2007.1, 2020.

\bibitem{LUNDERVOLD2019102}
Alexander~Selvikvåg Lundervold and Arvid Lundervold.
\newblock {An overview of deep learning in medical imaging focusing on MRI}.
\newblock {\em Zeitschrift f{\"{u}}r Medizinische Physik}, 29(2):102--127,
  2019.

\bibitem{Nasr2018ComprehensivePA}
Milad Nasr, R~Shokri, and Amir Houmansadr.
\newblock {Comprehensive Privacy Analysis of Deep Learning: Stand-alone and
  Federated Learning under Passive and Active White-box Inference Attacks}.
\newblock {\em ArXiv}, abs/1812.0, 2018.

\bibitem{Nasr2018MachineLW}
Milad Nasr, R~Shokri, and Amir Houmansadr.
\newblock {Machine Learning with Membership Privacy using Adversarial
  Regularization}.
\newblock {\em Proceedings of the 2018 ACM SIGSAC Conference on Computer and
  Communications Security}, 2018.

\bibitem{Nicolae2018AdversarialRT}
Maria-Irina Nicolae, Mathieu Sinn, Minh-Ngoc Tran, Beat Buesser, Ambrish Rawat,
  Martin Wistuba, Valentina Zantedeschi, Nathalie Baracaldo, Bryant Chen, Heiko
  Ludwig, Ian Molloy, and Ben Edwards.
\newblock {Adversarial Robustness Toolbox v1.0.0}.
\newblock {\em arXiv: Learning}, 2018.

\bibitem{NEURIPS2020_e6385d39}
Garima Pruthi, Frederick Liu, Satyen Kale, and Mukund Sundararajan.
\newblock {Estimating Training Data Influence by Tracing Gradient Descent}.
\newblock In H~Larochelle, M~Ranzato, R~Hadsell, M~F Balcan, and H~Lin,
  editors, {\em NeurIPS}, volume~33, pages 19920--19930. Curran Associates,
  Inc., 2020.

\bibitem{Rezaei2021OnTD}
Shahbaz Rezaei and Xin Liu.
\newblock {On the Difficulty of Membership Inference Attacks}.
\newblock {\em CVPR}, pages 7888--7896, 2021.

\bibitem{7424435}
Mauro Ribeiro, Katarina Grolinger, and Miriam A~M Capretz.
\newblock {MLaaS: Machine Learning as a Service}.
\newblock In {\em ICMLA}, pages 896--902, 2015.

\bibitem{Sablayrolles2019WhiteboxVB}
Alexandre Sablayrolles, Matthijs Douze, Cordelia Schmid, Yann Ollivier, and
  Hervé J{\'{e}}gou.
\newblock {White-box vs Black-box: Bayes Optimal Strategies for Membership
  Inference}.
\newblock In {\em ICML}, 2019.

\bibitem{Salem2019MLLeaksMA}
Ahmed Salem, Yang Zhang, Mathias Humbert, Mario Fritz, and Michael Backes.
\newblock {ML-Leaks: Model and Data Independent Membership Inference Attacks
  and Defenses on Machine Learning Models}.
\newblock {\em ArXiv}, abs/1806.0, 2019.

\bibitem{Schioppa2021ScalingUI}
Andrea Schioppa, Polina Zablotskaia, David Vilar, and Artem Sokolov.
\newblock {Scaling Up Influence Functions}.
\newblock {\em ArXiv}, abs/2112.0, 2021.

\bibitem{Shao_Skryagin_Stammer_Schramowski_Kersting_2021}
Xiaoting Shao, Arseny Skryagin, Wolfgang Stammer, Patrick Schramowski, and
  Kristian Kersting.
\newblock {Right for Better Reasons: Training Differentiable Models by
  Constraining their Influence Functions}.
\newblock {\em AAAI}, 35(11):9533--9540, 5 2021.

\bibitem{Shokri2017MembershipIA}
R~Shokri, Marco Stronati, Congzheng Song, and Vitaly Shmatikov.
\newblock {Membership Inference Attacks Against Machine Learning Models}.
\newblock {\em 2017 IEEE Symposium on Security and Privacy (SP)}, pages 3--18,
  2017.

\bibitem{Song2019PrivacyRO}
Liwei Song, R~Shokri, and Prateek Mittal.
\newblock {Privacy Risks of Securing Machine Learning Models against
  Adversarial Examples}.
\newblock {\em Proceedings of the 2019 ACM SIGSAC Conference on Computer and
  Communications Security}, 2019.

\bibitem{Taigman2014DeepFaceCT}
Yaniv Taigman, Ming Yang, Marc'Aurelio Ranzato, and Lior Wolf.
\newblock {DeepFace: Closing the Gap to Human-Level Performance in Face
  Verification}.
\newblock {\em CVPR}, pages 1701--1708, 2014.

\bibitem{Truex2018TowardsDM}
Stacey Truex, Ling Liu, Mehmet~Emre Gursoy, Lei Yu, and Wenqi Wei.
\newblock {Towards Demystifying Membership Inference Attacks}.
\newblock {\em ArXiv}, abs/1807.0, 2018.

\bibitem{Yeom2018PrivacyRI}
Samuel Yeom, Irene Giacomelli, Matt Fredrikson, and Somesh Jha.
\newblock {Privacy Risk in Machine Learning: Analyzing the Connection to
  Overfitting}.
\newblock {\em 2018 IEEE 31st Computer Security Foundations Symposium (CSF)},
  pages 268--282, 2018.

\end{thebibliography}
}
\clearpage
\appendix
\section{SIF algorithm}
\label{supp_sec:sif_algorithm}
\setcounter{algorithm}{0}
\renewcommand{\thealgorithm}{A\arabic{algorithm}}
Algorithm~\ref{alg:train_SIF} summarizes the fitting of our self-influence function (SIF) attack model $\mathcal{A}$. For every sample in the training set $\mathcal{D}_{mem}^{train}$ or $\mathcal{D}_{non-mem}^{train}$ (defined in Section~\ref{sec:attack_model_training_and_evaluation}), we collect the $I_{SIF}$ measure (Eq. (3)) together with a variable $m$ that indicates if the target model $h$ predicted the same class as the groundtruth label. These values are then used to calculate the parameters, $\tau_1$ and $\tau_2$, of the attack model $\mathcal{A}$ that is provided by Algorithm~\ref{alg:set_thresholds} (line \#25).

The procedure in Algorithm~\ref{alg:set_thresholds} aims to find an interval ($\tau_1$, $\tau_2$) that best encapsulates only the members, i.e., we want to have that most of the members' SIF values are inside ($\tau_1$, $\tau_2$) and most of the non-members' SIF values are outside this range. Since the SIF values distribution does not resemble a Gaussian (see Figure~\ref{fig:sif_hist}), we consider $1000$ samples distributed uniformly around both the members' minimum and maximum values (lines \#10-11). For every possible pair $\tau_1$, $\tau_2$ (\#lines 14-15) we calculate the balanced accuracy as defined in Eq.~\eqref{eq:balanced_accuracy}. The optimal threshold pair is selected based on a maximization of the balanced accuracy on the training set.

Lastly, Algorithm~\ref{alg:SIF_inference} shows the inference of our attack model $\mathcal{A}$. Given a target model $h$ and data sample $z=(x,y)$, we calculate the SIF value $s$ and query $h$ for its class prediction. If both conditions are met: (i) $s \in (\tau_1, \tau_2)$ and (ii) $y = \hat{y}$ (where $\hat{y}=h(x;\theta)$), then $\mathcal{A}$ predicts $z$ as a member. Otherwise, $z$ is predicted as a non-member.

\begin{algorithm*}[ht!]
\caption{Fitting self-influence function (SIF) attack}\label{alg:train_SIF}
\begin{algorithmic}[1]
\Require Training set $\big\{\mathcal{D}_{mem}^{train} \cup \mathcal{D}_{non-mem}^{train}\big\} \subset \mathcal{X} \times \mathcal{Y}$ 
\Require $h(x; \theta)$ Pre-trained target model with parameters $\theta$
\Ensure Attack model $\mathcal{A}(x, y; \tau_1, \tau_2)$ \Comment{A membership inference predictor}

\item[]

\State Initialize: $SIF_m$=[], $SIF_{nm}$=[]\Comment{SIF values}
\State Initialize: $M_m$=[], $M_{nm}$=[]\Comment{Whether the $h(x,\theta)$ class predictions matches the label}

\For{$z=(x, y)$ in $\mathcal{D}_{mem}^{train}$}
  \State $s \leftarrow I_{SIF}(z)$ \Comment{Eq.~\eqref{eq:SIF}}
  \State $\hat{y} \leftarrow h(x;\theta)$ \Comment{Query target model}
  \If{$\hat{y} == y$}
    \State $m\leftarrow1$
  \Else
    \State $m\leftarrow0$
  \EndIf
  \State $SIF_m$.append($s$)
  \State $M_m$.append($m$)
\EndFor

\For{$z=(x, y)$ in $\mathcal{D}_{non-mem}^{train}$}
  \State $s \leftarrow I_{SIF}(z)$
  \State $\hat{y} \leftarrow h(x;\theta)$
  \If{$\hat{y} == y$}
    \State $m\leftarrow1$
  \Else
    \State $m\leftarrow0$
  \EndIf
  \State $SIF_{nm}$.append($s$)
  \State $M_{nm}$.append($m$)
\EndFor

\State \textbf{set} $\tau_1, \tau_2$ := \textsc{setThresholds}($SIF_{m}$, $M_{m}$, $SIF_{nm}$, $M_{nm}$)
\end{algorithmic}
\end{algorithm*}

\begin{algorithm*}[ht!]
\caption{Setting $\tau_1$ and $\tau_2$ thresholds for attack model $\mathcal{A}$}\label{alg:set_thresholds}
\begin{algorithmic}[1]
\Procedure{setThresholds}{$SIF_{m}$, $M_{m}$, $SIF_{nm}$, $M_{nm}$}
\State $N_1 = |SIF_{m}|$ \Comment{$N_1$ is the total number of members}
\State $N_2 = |SIF_{nm}|$ \Comment{$N_2$ is the total number of non-members}
\State $best\_acc \leftarrow 0$
\State $best\_\tau_1 \leftarrow -\infty$
\State $best\_\tau_2 \leftarrow \infty$

\State $SIF_m^{min} \leftarrow min(SIF_{m})$ 
\State $SIF_m^{max} \leftarrow max(SIF_{m})$
\State $\delta \leftarrow SIF_m^{max} - SIF_m^{min}$

\item[]
\State $min\_arr := \text{linspace}(SIF_m^{min} - \frac{\delta}{2}, SIF_m^{min} + \frac{\delta}{2}, 1000)$ 
\State $max\_arr := \text{linspace}(SIF_m^{max} - \frac{\delta}{2}, SIF_m^{max} + \frac{\delta}{2}, 1000)$ 
\For{$i$ in $[1:1000]$}
  \For{$j$ in $[1:1000]$}
    \State $\tau_1 \leftarrow min\_arr[i]$
    \State $\tau_2 \leftarrow max\_arr[j]$
    \State Initialize: $\hat{y}_m$=[], $\hat{y}_{nm}$=[] \Comment{Set MI prediction vectors for members and non-members}
    \For{$k$ in $[1:N_1]$}
      \If{$\tau_1 < SIF_m[k]$ and $SIF_m[k] < \tau_2$ and $M_m[k] == 1$}
        \State $\hat{y}_m$.append($1$)
      \Else
        \State $\hat{y}_m$.append($0$)
      \EndIf
    \EndFor
    \For{$k$ in $[1:N_2]$}
      \If{$\tau_1 < SIF_{nm}[k]$ and $SIF_{nm}[k] < \tau_2$ and $M_{nm}[k] == 1$}
        \State $\hat{y}_{nm}$.append($1$)
      \Else
        \State $\hat{y}_{nm}$.append($0$)
      \EndIf
    \EndFor
    \State $acc \leftarrow$ Balanced Acc($\hat{y}_m$, $\hat{y}_{nm}$) \Comment{Eq.~\eqref{eq:balanced_accuracy}}
    \If{$acc > best\_acc$}
      \State $best\_acc \leftarrow acc$
      \State $best\_\tau_1 \leftarrow \tau_1$
      \State $best\_\tau_2 \leftarrow \tau_2$
    \EndIf
  \EndFor
\EndFor
\State \textbf{return} $best\_\tau_1$, $best\_\tau_2$
\EndProcedure
\end{algorithmic}
\end{algorithm*}

\begin{algorithm*}[ht!]
\caption{SIF inference}\label{alg:SIF_inference}
\begin{algorithmic}[1]
\Require $h(x; \theta)$ Pre-trained target model with parameters $\theta$
\Require $\mathcal{A}(x, y; \tau_1, \tau_2)$ Pre-trained attack model with parameters $\tau_1$ and $\tau_2$
\Require $z=(x,y)$ Data sample
\Ensure Membership inference prediction \Comment{$1$ for member and $0$ for non-member}
\item[]
\State $s \leftarrow I_{SIF}(z)$ \Comment{Eq.~\eqref{eq:SIF}}
\State $\hat{y} \leftarrow h(x;\theta)$ \Comment{Query target model}
\If{$\tau_1 < s$ and $s < \tau_2$ and $\hat{y} == y$}
  \State \textbf{return} $1$
\Else
  \State \textbf{return} $0$
\EndIf
\end{algorithmic}
\end{algorithm*}
\clearpage
\section{SIF and adaSIF calculation}
\label{supp_sec:efficient_sif_calculation}
\setcounter{equation}{0}
\renewcommand{\theequation}{B\arabic{equation}}
Here we explain in detail how we calculated the $I_{SIF}$ and $I_{adaSIF}$ values in Eq.~\eqref{eq:SIF} and Eq.~\eqref{eq:adaSIF}, respectively.
\subsection{SIF}
The vanilla SIF value is given by:
$$I_{SIF}(z) = -\nabla_{\theta}L(z, \hat{\theta})^TH_{\hat{\theta}}^{-1}\nabla_{\theta}L(z, \hat{\theta}).
$$
Since the Hessian $H_{\theta}$ and its inverse are not feasible to compute due to millions of parameters in deep neural networks (DNNs), we avoid their computation completely and follow the method shown in \cite{Koh17Understanding}. We approximate $I_{SIF}$ using Hessian vector products (HVPs):
\begin{equation}
\label{eq:I_sif_detailed}
I_{SIF}(z) = -\underbrace{H_{\hat{\theta}}^{-1}\nabla_{\theta}L(z, \hat{\theta})}_{s(z)} \cdot \underbrace{\nabla_{\theta}L(z, \hat{\theta})}_{grad_z}.
\end{equation}

Koh and Liang \cite{Koh17Understanding} employed this HVP and approximated $s(z)$ using stochastic estimation \cite{Agarwal2016SecondOS}, while iterating over data points from the training set. In our vanilla SIF case, we use their $s(z)$ approximation with one iteration since we consider the self-influence of a single data point. The $grad_z$ value is the gradient map from the loss to the image plane, and is calculated with a simple back-propagation pass.

\subsection{adaSIF}
Here we consider a scenario where the target model was trained with data augmentations. Let $z=(x,y)$ denote an original sample and $I$ be a random data augmentation operator sampled from the family of training augmentation distribution $\mathcal{T} \big(I\sim\mathcal{T}\big)$. We approximate $s(z)$ and $grad_z$ in Eq.~\eqref{eq:I_sif_detailed} by taking their expected value over these transformations. Formally, we calculate:
\begin{equation}
\label{eq:I_adasif_detailed}
\begin{split}
I_{adaSIF}(z) &\eqdef  -\mathbb{E}_{I\sim\mathcal{T}}\big[s(z)\big] \cdot \mathbb{E}_{I\sim\mathcal{T}}\big[grad_z\big] \\
           &= -\underbrace{\mathbb{E}_{I\sim\mathcal{T}}\Big[H_{\hat{\theta}}^{-1}\nabla_{\theta}L\big(I(x), y, \hat{\theta}\big)\Big]}_{(i)} \cdot \underbrace{\mathbb{E}_{I\sim\mathcal{T}}\Big[\nabla_{\theta}L\big(I(x), y, \hat{\theta}\big)\Big]}_{(ii)}.
\end{split}
\end{equation}

For approximating $(i)$ we employ the same stochastic estimation as used by Koh and Liang, but instead of iterating over different data points, we iterate over a set of image transformations. $(ii)$ is calculated by averaging gradient maps of $128$ different image transformations $I(x)$.

\section{Hardware Setup}
\label{supp_sec:hardware_setup}
All the target models and attack models were trained and evaluated on a machine with a GPU of type NVIDIA GeForce RTX 2080 Ti, which has 11 GB of VRAM. For training the target models we utilized 4 threads of Intel Xeon Silver 4114 CPU. The target models were evaluated using a single CPU core.
All the attack models' fitting and inference were performed using a single GPU and a single CPU core.
\clearpage

\section{Accuracy of target models}
\label{supp_sec:accuracy_of_target_models}
\setcounter{table}{0}
\renewcommand{\thetable}{D\arabic{table}}
Table~\ref{tab:target_models_accuracies} and Table~\ref{tab:target_models_accuracies_aug} report the \textit{training}, \textit{validation}, and \textit{test} accuracies of target models trained without and with data augmentations, respectively, as defined in Section~\ref{sec:target_model}. Notice that Tiny ImageNet was not trained on $\mathcal{M}$-1 since the dataset has 200 labels whereas the smallest target model has only $100$ data points for training. All target models exhibit sufficient test accuracy for a meaningful MI analysis.
\begin{table}[ht!]
\caption{The \textit{training}, \textit{validation}, and \textit{test} accuracies [\%] for all the target models used in the paper, that were trained without data augmentations.}
\centering
\resizebox{1\columnwidth}{!}{
\begin{tabular}{cc|ccc|ccc|ccc}
\toprule
\multicolumn{2}{l|}{\multirow{2}{*}{Target Models}} & \multicolumn{3}{c|}{CIFAR-10} & \multicolumn{3}{c|}{CIFAR-100} & \multicolumn{3}{c}{Tiny ImageNet} \\
& & Train & Val & Test & Train & Val & Test & Train & Val & Test \\
\hline
\multirow{7}{*}{AlexNet} & $\mathcal{M}$-1 & 21.00  & 19.80 & 19.95 & 1.00   & 2.00  & 1.74  & - & - & - \\
                         & $\mathcal{M}$-2 & 54.30  & 35.56 & 33.27 & 100.00 & 6.00  & 5.75  & 0.60   & 1.20  & 0.96 \\
                         & $\mathcal{M}$-3 & 100.00 & 50.52 & 51.77 & 36.88 & 13.00  & 12.59 & 20.02  & 4.84  & 4.12 \\
                         & $\mathcal{M}$-4 & 100.00 & 61.72 & 60.29 & 99.98 & 17.20  & 18.38 & 10.99  & 4.32  & 3.93 \\
                         & $\mathcal{M}$-5 & 100.00 & 65.68 & 64.33 & 99.99 & 27.24  & 26.60 & 100.00 & 8.44  & 7.30 \\
                         & $\mathcal{M}$-6 & 100.00 & 67.56 & 67.70 & 99.98 & 26.56  & 27.32 & 23.02  & 11.22 & 10.46 \\
                         & $\mathcal{M}$-7 & 100.00 & 71.28 & 70.55 & 99.96 & 34.20  & 33.40 & 22.42  & 14.54 & 13.76 \\
\hline
\multirow{7}{*}{ResNet18}& $\mathcal{M}$-1 & 100.00 & 19.32 & 19.49 & 100.00 & 3.48  & 3.35  & - & - & - \\
                         & $\mathcal{M}$-2 & 100.00 & 39.00 & 38.62 & 100.00 & 11.88 & 10.91 & 99.60  & 2.80  & 2.87 \\
                         & $\mathcal{M}$-3 & 100.00 & 57.24 & 56.94 & 100.00 & 22.84 & 22.80 & 99.98  & 8.56  & 8.51 \\
                         & $\mathcal{M}$-4 & 100.00 & 67.88 & 65.00 & 99.99  & 31.24 & 30.19 & 100.00 & 14.38 & 14.40 \\
                         & $\mathcal{M}$-5 & 100.00 & 76.20 & 74.18 & 100.00 & 38.32 & 40.10 & 100.00 & 19.90 & 19.42 \\
                         & $\mathcal{M}$-6 & 100.00 & 76.04 & 74.42 & 100.00 & 48.20 & 47.12 & 100.00 & 23.38 & 23.04 \\
                         & $\mathcal{M}$-7 & 100.00 & 76.96 & 76.30 & 99.99  & 48.56 & 47.88 & 100.00 & 25.84 & 25.07 \\
\hline
\multirow{7}{*}{DenseNet}& $\mathcal{M}$-1 & 100.00 & 24.80 & 24.82 & 100.00 & 3.24  & 2.86  & - & - & - \\
                         & $\mathcal{M}$-2 & 100.00 & 45.84 & 45.86 & 99.80  & 11.40 & 10.74 & 99.20 & 3.24  & 3.04 \\
                         & $\mathcal{M}$-3 & 100.00 & 65.84 & 64.88 & 99.64  & 27.48 & 25.96 & 42.74 & 11.14 & 10.72 \\
                         & $\mathcal{M}$-4 & 100.00 & 75.16 & 74.28 & 99.99  & 36.24 & 36.11 & 34.02 & 16.20 & 15.18 \\
                         & $\mathcal{M}$-5 & 100.00 & 77.80 & 77.51 & 97.27  & 41.32 & 40.48 & 27.79 & 19.38 & 19.01 \\
                         & $\mathcal{M}$-6 & 100.00 & 81.08 & 79.92 & 94.24 & 45.64 & 44.70 & 46.77 & 23.46 & 23.27 \\
                         & $\mathcal{M}$-7 & 100.00 & 82.96 & 81.97 & 81.71 & 47.16 & 46.30 & 41.97 & 24.90 & 25.06 \\
\bottomrule
\end{tabular}
}
\label{tab:target_models_accuracies}
\vspace{-0.18in}
\end{table}
\clearpage

\begin{table}[ht!]
\caption{The \textit{training}, \textit{validation}, and \textit{test} accuracies [\%] for all the target models used in the paper, that were trained with data augmentations.}
\centering
\resizebox{1\columnwidth}{!}{
\begin{tabular}{cc|ccc|ccc|ccc}
\toprule
\multicolumn{2}{l|}{\multirow{2}{*}{Target Models}} & \multicolumn{3}{c|}{CIFAR-10} & \multicolumn{3}{c|}{CIFAR-100} & \multicolumn{3}{c}{Tiny ImageNet} \\
& & Train & Val & Test & Train & Val & Test & Train & Val & Test \\
\hline
\multirow{7}{*}{AlexNet} & $\mathcal{M}$-1 & 39.00 & 25.76 & 26.43 & 15.00 & 3.12  & 2.23  & - & - & - \\ 
                         & $\mathcal{M}$-2 & 99.40 & 44.28 & 42.06 & 97.60 & 8.96  & 7.71  & 98.60  & 2.86 & 2.77 \\
                         & $\mathcal{M}$-3 & 99.60 & 69.68 & 67.64 & 41.42 & 14.80 & 13.46 & 100.00 & 9.50 & 9.11 \\
                         & $\mathcal{M}$-4 & 99.76 & 74.56 & 73.70 & 49.00 & 22.36 & 21.45 & 13.24  & 6.70 & 6.51 \\
                         & $\mathcal{M}$-5 & 99.86 & 78.16 & 77.14 & 99.98 & 28.60 & 28.54 & 14.47  & 9.18 & 8.82 \\
                         & $\mathcal{M}$-6 & 99.91 & 80.40 & 79.08 & 99.96 & 32.88 & 31.89 & 99.99  & 13.76 & 12.49 \\
                         & $\mathcal{M}$-7 & 99.44 & 79.92 & 80.04 & 99.96 & 36.72 & 36.46 & 99.90  & 17.74 & 16.82 \\
\hline
\multirow{7}{*}{ResNet18}& $\mathcal{M}$-1 & 100.00 & 21.96 & 22.66 & 100.00 & 3.80  & 3.97  & - & - & - \\
                         & $\mathcal{M}$-2 & 99.80  & 42.16 & 41.15 & 100.00 & 12.80 & 12.07 & 99.80  & 3.54 & 2.92 \\
                         & $\mathcal{M}$-3 & 100.00 & 69.16 & 67.83 & 100.00 & 31.28 & 31.42 & 100.00 & 15.50 & 15.27 \\
                         & $\mathcal{M}$-4 & 100.00 & 79.08 & 76.91 & 99.99  & 48.96 & 48.27 & 100.00 & 26.40 & 24.87 \\
                         & $\mathcal{M}$-5 & 100.00 & 86.56 & 85.89 & 100.00 & 56.52 & 57.09 & 99.99  & 30.26 & 30.67 \\
                         & $\mathcal{M}$-6 & 100.00 & 91.60 & 90.25 & 100.00 & 64.44 & 62.89 & 99.99  & 35.18 & 35.51 \\
                         & $\mathcal{M}$-7 & 100.00 & 91.84 & 90.29 & 100.00 & 63.72 & 63.62 & 99.99  & 39.40 & 38.95 \\
\hline
\multirow{7}{*}{DenseNet}& $\mathcal{M}$-1 & 100.00 & 24.52 & 24.23 & 100.00 & 4.36  & 4.17  & - & - & - \\
                         & $\mathcal{M}$-2 & 100.00 & 51.48 & 49.55 & 97.50  & 10.72 & 10.29 & 98.40 & 3.32 & 3.35 \\
                         & $\mathcal{M}$-3 & 99.74  & 73.76 & 71.99 & 99.38  & 34.20 & 34.57 & 88.92 & 13.06 & 12.23 \\
                         & $\mathcal{M}$-4 & 99.62  & 80.28 & 79.15 & 88.64  & 44.56 & 43.09 & 71.37 & 22.02 & 20.42 \\
                         & $\mathcal{M}$-5 & 99.63  & 84.88 & 84.09 & 93.07  & 50.12 & 48.76 & 57.25 & 26.16 & 25.70 \\
                         & $\mathcal{M}$-6 & 99.47  & 87.72 & 85.25 & 88.41  & 54.24 & 52.40 & 59.69 & 30.88 & 29.89 \\
                         & $\mathcal{M}$-7 & 99.36  & 87.32 & 85.96 & 79.97  & 55.56 & 54.71 & 63.91 & 33.86 & 33.10 \\
\bottomrule
\end{tabular}
}
\label{tab:target_models_accuracies_aug}
\vspace{-0.18in}
\end{table}
\clearpage
\section{Comparison of MI attacks}
\label{supp_sec:comparison_of_mi_attacks}
\setcounter{figure}{0}
\renewcommand{\thefigure}{E\arabic{figure}}
Here we continue the MI attack comparison from Section~\ref{sec:comparison_of_mi_attacks}, and include other architectures. Figure~\ref{fig:alexnet_densenet_attack_scores_small} presents the balanced accuracy on target models trained on AlexNet and DenseNet. We observe that in most cases SIF performs on par with current state-of-the-art (SOTA). A new SOTA is achieved for CIFAR-10 trained on DenseNet (Figure~\ref{fig:alexnet_densenet_attack_scores_small}(d)).

\begin{figure}[ht!]
\centering
\includegraphics[width=1\linewidth]{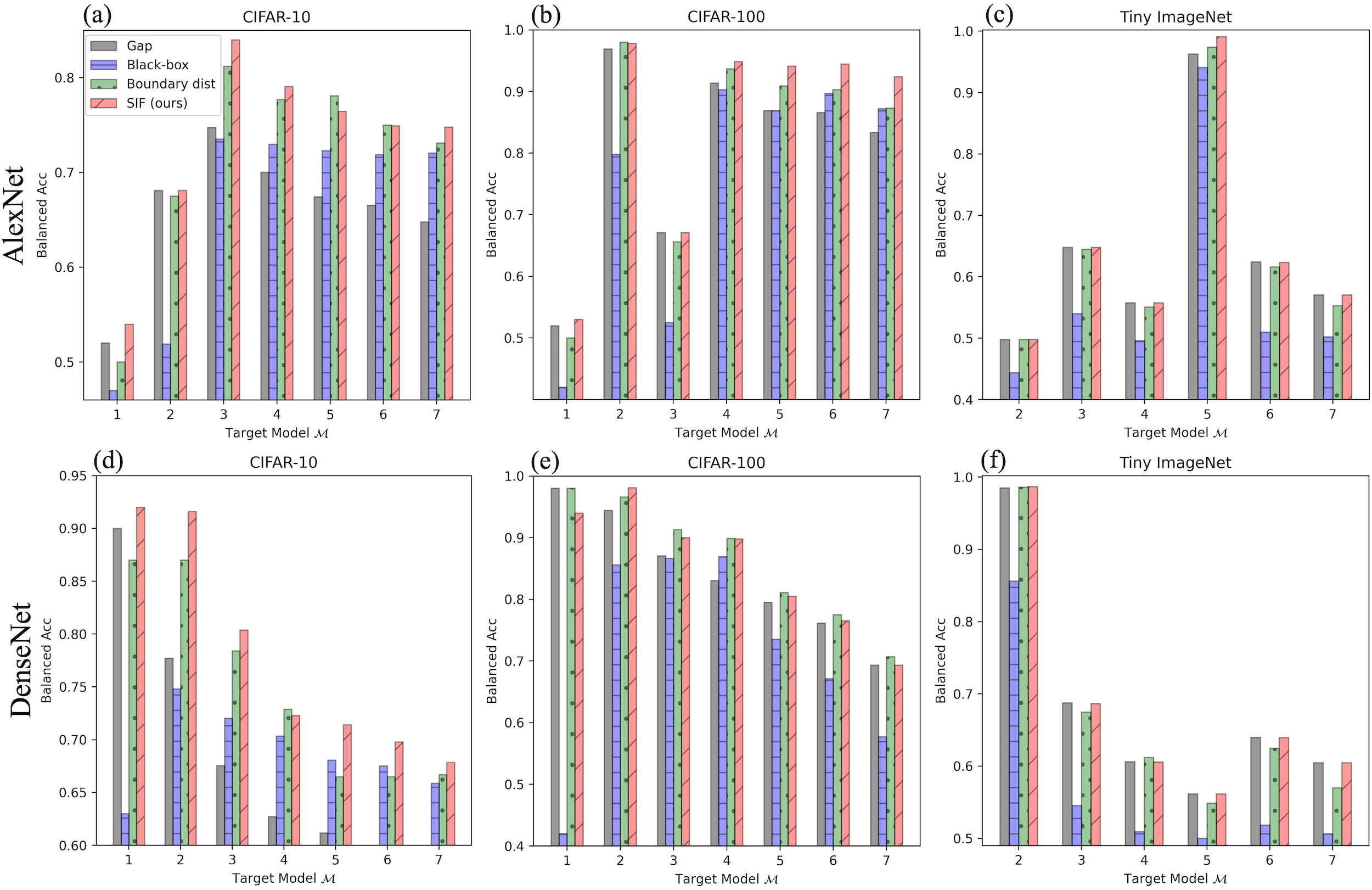}
\vspace{-0.24in}
\caption{Comparison of our SIF attack with some baseline MI attacks: Gap, Black-box, and Boundary distance. The top and bottom rows show target models trained using AlexNet and DenseNet architectures, respectively. The x-axis indicates the attacked target model and the y-axis shows the balanced attack accuracy (Eq.~\eqref{eq:balanced_accuracy}).}
\label{fig:alexnet_densenet_attack_scores_small}
\vspace{-0.18in}
\end{figure}

\clearpage
\section{Precision and recall}
\label{supp_sec:precision_and_recall}
\setcounter{table}{0}
\renewcommand{\thetable}{F\arabic{table}}
Tables~\ref{tab:precision_and_recall_cifar10}-\ref{tab:precision_and_recall_tiny_imagenet} show all attack models' precision, recall, and accuracy metrics for members and non-members, on target models trained with ResNet18, for CIFAR-10, CIFAR-100, and Tiny ImageNet, respectively. In addition to the excellent balanced accuracy we reported in Section~\ref{sec:comparison_of_mi_attacks}, SIF also achieves superb precision for members and recall for non-members, surpassing previous SOTA by a large margin for the majority of cases. These results demonstrate that our SIF attack does not suffer from the high False Alarm Rate (FAR) observed in many other MI inference attacks \cite{Rezaei2021OnTD}, making it a very reliable method for detecting training set samples. We Also observe prefect recall ($\sim 1.0$) for members, matching our baselines and the results from \cite{Shokri2017MembershipIA,Yeom2018PrivacyRI}.

\begin{table}[h!]
\caption{Accuracy, precision, and recall for members and non-members. Target models were trained on CIFAR-10 with ResNet18.}
\centering
\resizebox{1\columnwidth}{!}{
\begin{tabular}{cc|ccc|ccc|c}
\toprule
\multirow{2}{*}{\parbox{1cm}{Target model}} & \multirow{2}{*}{Attack model} & \multicolumn{3}{c|}{Member} & \multicolumn{3}{c|}{Non-member} & \multirow{2}{*}{Balanced Acc} \\
& & Acc & Precision & Recall & Acc & Precision & Recall & \\
\hline
\multirow{4}{*}{$\mathcal{M}$-1} & Gap           & \textbf{1.00} & 0.82 & \textbf{1.00} & 0.78 & \textbf{1.00} & 0.78 & 0.89 \\
                                 & Black-box     & 0.60 & 0.54 & 0.60 & 0.48 & 0.55 & 0.48 & 0.54 \\
                                 & Boundary dist & 0.98 & 0.93 & 0.98 & 0.92 & 0.98 & 0.92 & 0.95 \\
                                 & SIF (ours)    & \textbf{1.00} & \textbf{0.98} & \textbf{1.00} & \textbf{0.98} & \textbf{1.00} & \textbf{0.98} & \textbf{0.99} \\
\hline
\multirow{4}{*}{$\mathcal{M}$-2} & Gap           & \textbf{1.00} & 0.72 & \textbf{1.00} & 0.61 & \textbf{1.00} & 0.61 & 0.81 \\
                                 & Black-box     & \textbf{1.00} & 0.72 & \textbf{1.00} & 0.62 & \textbf{1.00} & 0.62 & 0.81 \\
                                 & Boundary dist & 0.99 & 0.84 & 0.99 & 0.81 & 0.99 & 0.81 & 0.90 \\
                                 & SIF (ours)    & \textbf{1.00} & \textbf{0.91} & \textbf{1.00} & \textbf{0.91} & \textbf{1.00} & \textbf{0.91} & \textbf{0.95} \\
\hline
\multirow{4}{*}{$\mathcal{M}$-3} & Gap           & \textbf{1.00}	& 0.63 & \textbf{1.00} & 0.41 & \textbf{1.00} & 0.41 & 0.71 \\
                                 & Black-box     & \textbf{1.00}	& 0.75 & \textbf{1.00} & 0.67 & \textbf{1.00} & 0.67 & 0.83 \\
                                 & Boundary dist & 0.95	& 0.79 & 0.95 & 0.75 & 0.93	& 0.75 & 0.85 \\
                                 & SIF (ours)    & \textbf{1.00}	& \textbf{0.85} & \textbf{1.00} & \textbf{0.82} & \textbf{1.00}	& \textbf{0.82} & \textbf{0.91} \\
\hline
\multirow{4}{*}{$\mathcal{M}$-4} & Gap           & \textbf{1.00} & 0.61 & \textbf{1.00} & 0.36 & \textbf{1.00} & 0.36 & 0.68 \\
                                 & Black-box     & 0.99	& 0.74 & 0.99 & 0.65 & 0.98	& 0.65 & 0.82 \\
                                 & Boundary dist & 0.91	& 0.74 & 0.91 & 0.68 & 0.89	& 0.68 & 0.80 \\
                                 & SIF (ours)    & 0.99	& \textbf{0.80} & 0.99 & \textbf{0.75} & 0.99	& \textbf{0.75} & \textbf{0.87} \\
\hline
\multirow{4}{*}{$\mathcal{M}$-5} & Gap           & \textbf{1.00}	& 0.57 & \textbf{1.00} & 0.25 & \textbf{1.00}	& 0.25 & 0.63 \\
                                 & Black-box     & \textbf{1.00}	& 0.67 & \textbf{1.00} & 0.51 & \textbf{1.00}	& 0.51 & 0.76 \\
                                 & Boundary dist & 0.95	& 0.70 & 0.95 & 0.59 & 0.92	& 0.59 & 0.77 \\
                                 & SIF (ours)    & 0.99	& \textbf{0.73} & 0.99 & \textbf{0.64} & 0.98	& \textbf{0.64} & \textbf{0.81} \\
\hline
\multirow{4}{*}{$\mathcal{M}$-6} & Gap           & \textbf{1.00}	& 0.57 & \textbf{1.00} & 0.24 & \textbf{1.00}	& 0.24 & 0.62 \\
                                 & Black-box     & 0.89	& 0.71 & 0.89 & \textbf{0.63} & 0.85	& \textbf{0.63} & 0.76 \\
                                 & Boundary dist & 0.97	& 0.66 & 0.97 & 0.50 & 0.94	& 0.50 & 0.74 \\
                                 & SIF (ours)    & 0.98	& \textbf{0.72} & 0.98 & 0.62 & 0.96	& 0.62 & \textbf{0.80} \\
\hline
\multirow{4}{*}{$\mathcal{M}$-7} & Gap           & \textbf{1.00}	& 0.57 & \textbf{1.00} & 0.23 & \textbf{1.00}	& 0.23 & 0.62 \\
                                 & Black-box     & \textbf{1.00}	& \textbf{0.70} & \textbf{1.00} & 0.58 & \textbf{1.00}	& 0.58 & \textbf{0.79} \\
                                 & Boundary dist & 0.91	& \textbf{0.70} & 0.91 & \textbf{0.62} & 0.87	& \textbf{0.62} & 0.76 \\
                                 & SIF (ours)    & \textbf{1.00}	& 0.69 & \textbf{1.00} & 0.55 & \textbf{1.00}	& 0.55 & 0.78 \\
\bottomrule
\end{tabular}
}
\label{tab:precision_and_recall_cifar10}
\vspace{-0.18in}
\end{table}
\clearpage
\begin{table}[h!]
\caption{Accuracy, precision, and recall for members and non-members. Target models were trained on CIFAR-100 with ResNet18.}
\centering
\resizebox{1\columnwidth}{!}{
\begin{tabular}{cc|ccc|ccc|c}
\toprule
\multirow{2}{*}{\parbox{1cm}{Target model}} & \multirow{2}{*}{Attack model} & \multicolumn{3}{c|}{Member} & \multicolumn{3}{c|}{Non-member} & \multirow{2}{*}{Balanced Acc} \\
& & Acc & Precision & Recall & Acc & Precision & Recall & \\
\hline
\multirow{4}{*}{$\mathcal{M}$-1} & Gap           & \textbf{1.00}	& \textbf{1.00} & \textbf{1.00} & \textbf{1.00} & \textbf{1.00}	& \textbf{1.00} & \textbf{1.00} \\
                                 & Black-box     & 0.14	& 0.39 & 0.14 & 0.78 & 0.48	& 0.78 & 0.46 \\
                                 & Boundary dist & \textbf{1.00}	& \textbf{1.00} & \textbf{1.00} & \textbf{1.00} & \textbf{1.00}	& \textbf{1.00} & \textbf{1.00} \\
                                 & SIF (ours)    & \textbf{1.00}	& \textbf{1.00} & \textbf{1.00} & \textbf{1.00} & \textbf{1.00}	& \textbf{1.00} & \textbf{1.00} \\
\hline
\multirow{4}{*}{$\mathcal{M}$-2} & Gap           & \textbf{1.00}	& 0.87 & \textbf{1.00} & 0.85 & \textbf{1.00}	& 0.85 & 0.92 \\
                                 & Black-box     & \textbf{1.00}	& 0.87 & \textbf{1.00} & 0.85 & \textbf{1.00}	& 0.85 & 0.93 \\
                                 & Boundary dist & \textbf{1.00}	& 0.96 & \textbf{1.00} & 0.95 & \textbf{1.00}	& 0.95 & 0.98 \\
                                 & SIF (ours)    & \textbf{1.00}	& \textbf{0.99} & \textbf{1.00} & \textbf{0.99} & \textbf{1.00}	& \textbf{0.99} & \textbf{1.00} \\
\hline
\multirow{4}{*}{$\mathcal{M}$-3} & Gap           & \textbf{1.00}	& 0.81 & \textbf{1.00} & 0.76 & \textbf{1.00}	& 0.76 & 0.88\\
                                 & Black-box     & \textbf{1.00}	& 0.94 & \textbf{1.00} & 0.93 & \textbf{1.00}	& 0.93 & 0.97 \\
                                 & Boundary dist & 0.99	& 0.94 & 0.99 & 0.94 & 0.99	& 0.94 & 0.96 \\
                                 & SIF (ours)    & \textbf{1.00}	& \textbf{0.98} & \textbf{1.00} & \textbf{0.98} & \textbf{1.00}	& \textbf{0.98} & \textbf{0.99} \\
\hline
\multirow{4}{*}{$\mathcal{M}$-4} & Gap           & \textbf{1.00}	& 0.76 & 1.00 & 0.69 & \textbf{1.00}	& 0.69 & 0.85 \\
                                 & Black-box     & \textbf{1.00}	& 0.92 & 1.00 & 0.91 & \textbf{1.00}	& 0.91 & 0.96 \\
                                 & Boundary dist & \textbf{1.00}	& 0.91 & 1.00 & 0.90 & \textbf{1.00}	& 0.90 & 0.95 \\
                                 & SIF (ours)    & \textbf{1.00}	& \textbf{0.96} & 1.00 & \textbf{0.96} & \textbf{1.00}	& \textbf{0.96} & \textbf{0.98} \\
\hline
\multirow{4}{*}{$\mathcal{M}$-5} & Gap           & \textbf{1.00}	& 0.71 & \textbf{1.00} & 0.60 & \textbf{1.00}	& 0.60 & 0.80 \\
                                 & Black-box     & \textbf{1.00}	& 0.91 & \textbf{1.00} & 0.91 & \textbf{1.00}	& 0.91 & 0.95 \\
                                 & Boundary dist & 0.97	& 0.88 & 0.97 & 0.86 & 0.97	& 0.86 & 0.92 \\
                                 & SIF (ours)    & \textbf{1.00}	& \textbf{0.96} & \textbf{1.00} & \textbf{0.95} & \textbf{1.00}	& \textbf{0.95} & \textbf{0.98} \\
\hline
\multirow{4}{*}{$\mathcal{M}$-6} & Gap           & \textbf{1.00}	& 0.68 & \textbf{1.00} & 0.53 & \textbf{1.00}	& 0.53 & 0.77 \\
                                 & Black-box     & 0.99	& 0.90 & 0.99 & 0.89 & 0.99	& 0.89 & 0.94 \\
                                 & Boundary dist & 0.99	& 0.86 & 0.99 & 0.84 & 0.98	& 0.84 & 0.91 \\
                                 & SIF (ours)    & \textbf{1.00}	& \textbf{0.94} & \textbf{1.00} & \textbf{0.93} & \textbf{1.00}	& \textbf{0.93} & \textbf{0.97} \\
\hline
\multirow{4}{*}{$\mathcal{M}$-7} & Gap           & \textbf{1.00}	& 0.68 & \textbf{1.00} & 0.52 & \textbf{1.00} & 0.52 & 0.76 \\
                                 & Black-box     & 0.99	& 0.88 & 0.99 & 0.86 & 0.99	& 0.86 & 0.93 \\
                                 & Boundary dist & 0.98	& 0.83 & 0.98 & 0.80 & 0.97	& 0.80 & 0.89 \\
                                 & SIF (ours)    & \textbf{1.00}	& \textbf{0.91} & \textbf{1.00} & \textbf{0.90} & \textbf{1.00}	& \textbf{0.90} & \textbf{0.95} \\
\bottomrule
\end{tabular}
}
\label{tab:precision_and_recall_cifar100}
\vspace{-0.18in}
\end{table}
\clearpage
\begin{table}[h!]
\caption{Accuracy, precision, and recall for members and non-members. Target models were trained on Tiny ImageNet with ResNet18.}
\centering
\resizebox{1\columnwidth}{!}{
\begin{tabular}{cc|ccc|ccc|c}
\toprule
\multirow{2}{*}{\parbox{1cm}{Target model}} & \multirow{2}{*}{Attack model} & \multicolumn{3}{c|}{Member} & \multicolumn{3}{c|}{Non-member} & \multirow{2}{*}{Balanced Acc} \\
& & Acc & Precision & Recall & Acc & Precision & Recall & \\
\hline
\multirow{4}{*}{$\mathcal{M}$-2} & Gap           & \textbf{1.00}	& 0.96 & \textbf{1.00} & 0.96 & \textbf{1.00} & 0.96 & 0.98 \\
                                 & Black-box     & 0.94	& 0.92 & 0.94 & 0.91 & 0.94	& 0.91 & 0.93 \\
                                 & Boundary dist & 0.99	& \textbf{0.98} & 0.99 & \textbf{0.98} & 0.99	& \textbf{0.98} & 0.98 \\
                                 & SIF (ours)    & 0.99	& \textbf{0.98} & 0.99 & \textbf{0.98} & 0.99	& \textbf{0.98} & \textbf{0.99} \\
\hline
\multirow{4}{*}{$\mathcal{M}$-3} & Gap           & \textbf{1.00}	& 0.91 & \textbf{1.00} & 0.91 & \textbf{1.00}	& 0.91 & 0.95 \\
                                 & Black-box     & \textbf{1.00}	& 0.96 & \textbf{1.00} & 0.96 & \textbf{1.00}	& 0.96 & 0.98 \\
                                 & Boundary dist & \textbf{1.00}	& 0.98 & \textbf{1.00} & 0.98 & \textbf{1.00}	& 0.98 & 0.99 \\
                                 & SIF (ours)    & \textbf{1.00}	& \textbf{0.99} & \textbf{1.00} & \textbf{0.99} & \textbf{1.00}	& \textbf{0.99} & \textbf{1.00} \\
\hline
\multirow{4}{*}{$\mathcal{M}$-4} & Gap           & \textbf{1.00}	& 0.87 & \textbf{1.00} & 0.86 & \textbf{1.00} & 0.86 & 0.93 \\
                                 & Black-box     & \textbf{1.00}	& 0.98 & \textbf{1.00} & 0.98 & \textbf{1.00}	& 0.98 & \textbf{0.99} \\
                                 & Boundary dist & \textbf{1.00}	& 0.97 & \textbf{1.00} & 0.96 & \textbf{1.00}	& 0.96 & 0.98 \\
                                 & SIF (ours)    & \textbf{1.00}	& \textbf{0.99} & \textbf{1.00} & \textbf{0.99} & \textbf{1.00}	& \textbf{0.99} & \textbf{0.99} \\
\hline
\multirow{4}{*}{$\mathcal{M}$-5} & Gap           & \textbf{1.00}	& 0.84 & \textbf{1.00} & 0.81 & \textbf{1.00}	& 0.81 & 0.90 \\
                                 & Black-box     & \textbf{1.00}	& 0.97 & \textbf{1.00} & 0.97 & \textbf{1.00}	& 0.97 & 0.99 \\
                                 & Boundary dist & 0.99	& 0.96 & 0.99 & 0.96 & 0.99	& 0.96 & 0.98 \\
                                 & SIF (ours)    & \textbf{1.00}	& \textbf{0.99} & \textbf{1.00} & \textbf{0.99} & \textbf{1.00}	& \textbf{0.99} & \textbf{1.00} \\
\hline
\multirow{4}{*}{$\mathcal{M}$-6} & Gap           & \textbf{1.00}	& 0.82 & \textbf{1.00} & 0.78 & \textbf{1.00}	& 0.78 & 0.89 \\
                                 & Black-box     & \textbf{1.00}	& 0.95 & \textbf{1.00} & 0.95 & \textbf{1.00}	& 0.95 & 0.97 \\
                                 & Boundary dist & 0.99	& 0.95 & 0.99 & 0.94 & 0.99	& 0.94 & 0.97 \\
                                 & SIF (ours)    & \textbf{1.00}	& \textbf{0.97} & \textbf{1.00} & \textbf{0.97} & \textbf{1.00}	& \textbf{0.97} & \textbf{0.98} \\
\hline
\multirow{4}{*}{$\mathcal{M}$-7} & Gap           & \textbf{1.00}	& 0.80 & \textbf{1.00} & 0.75 & \textbf{1.00}	& 0.75 & 0.88 \\
                                 & Black-box     & 0.99	& \textbf{0.96} & 0.99 & \textbf{0.96} & 0.99	& \textbf{0.96} & \textbf{0.98} \\
                                 & Boundary dist & \textbf{1.00}	& 0.95 & \textbf{1.00} & 0.95 & \textbf{1.00}	& 0.95 & 0.97 \\
                                 & SIF (ours)    & \textbf{1.00}	& 0.93 & \textbf{1.00} & 0.93 & \textbf{1.00}	& 0.93 & 0.96 \\
\bottomrule
\end{tabular}
}
\label{tab:precision_and_recall_tiny_imagenet}
\vspace{-0.18in}
\end{table}
\clearpage

\section{Naive SIF ensemble}
\label{supp_sec:naive_sif_ensemble}
\setcounter{equation}{0}
\setcounter{table}{0}
\renewcommand{\theequation}{G\arabic{equation}}
\renewcommand{\thetable}{G\arabic{table}}
In Section~\ref{sec:adaptive_attack_to_augmentation} we propose to use a naive ensemble of SIF measures (named "avgSIF") to attack target models that are trained with data augmentation. Here we formally define avgSIF and compare its results to adaSIF. Let $z=(x,y)$ denote an original sample and $I$ be a random data augmentation operator sampled from the family of training augmentation distribution $\mathcal{T} \big(I\sim\mathcal{T}\big)$. Then we define the naive ensemble of SIF measures of $z$ as:
\begin{equation}
\label{eq:avgSIF}
\begin{split}
I_{avgSIF}(z) & \eqdef \mathbb{E}_{I\sim\mathcal{T}}\Big[I_{SIF}\big(I(x), y\big)\Big] \\
& = \mathbb{E}_{I\sim\mathcal{T}}\Big[-\nabla_{\theta}L\big(I(x), y, \hat{\theta}\big)^TH_{\hat{\theta}}^{-1}\nabla_{\theta}L\big(I(x), y, \hat{\theta}\big)\Big].
\end{split}
\end{equation}

The above term calculates $I_{SIF}$ scores (Eq.~\eqref{eq:SIF}) for $8$ different transformations of the input image $x$, and averages them to get the $I_{avgSIF}$ measure. The fitting and inference of the avgSIF attack
are done similarly to the vanilla SIF attack (see Section~\ref{supp_sec:sif_algorithm} for pseudo codes).

Table~\ref{tab:adasif_vs_avgsif} shows the accuracy, precision, and recall metrics for members and non-members, calculated for avgSIF and adaSIF, for target models trained on ResNet18. We observe that adaSIF outperforms avgSIF, achieving a higher balanced accuracy for the vast majority of the target models. In addition, adaSIF maintains a higher precision for the members which translates to a lower FAR (False Alarm Rate). Therefore, adaSIF was chosen for evaluating MI with data augmentation in Section~\ref{sec:data_augmentation_adaptive_attack}.
\begin{table}[ht!]
\caption{Comparison between MI attack performances of adaSIF and avgSIF. adaSIF is marginally better than avgSIF. We boldface the best member's precision and balanced accuracy.}
\hspace*{-0.29cm}
\centering
\resizebox{1\columnwidth}{!}{
\begin{tabular}{ccc|ccc|ccc|c}
\toprule
\multirow{2}{*}{Dataset} & \multirow{2}{*}{\parbox{1cm}{Target model}} & \multirow{2}{*}{Attack model} & \multicolumn{3}{c|}{Member} & \multicolumn{3}{c|}{Non-member} & \multirow{2}{*}{Balanced Acc} \\
& & & Acc & Precision & Recall & Acc & Precision & Recall & \\
\hline
\multirow{14}{*}{CIFAR-10} & \multirow{2}{*}{$\mathcal{M}$-1} & adaSIF & 0.940 & \textbf{0.825} & 0.940 & 0.800 & 0.930 & 0.800 & \textbf{0.870} \\
                                                            & & avgSIF & 0.960 & 0.814 & 0.960 & 0.780 & 0.951 & 0.780 & \textbf{0.870} \\
\cline{2-10}
                           & \multirow{2}{*}{$\mathcal{M}$-2} & adaSIF & 1.000 & \textbf{0.808} & 1.000 & 0.762 & 1.000 & 0.762 & \textbf{0.881} \\
                                                            & & avgSIF & 0.998 & 0.800 & 0.998 & 0.750 & 0.997 & 0.750 & 0.874 \\
\cline{2-10}
                           & \multirow{2}{*}{$\mathcal{M}$-3} & adaSIF & 0.993 & \textbf{0.761} & 0.993 & 0.688 & 0.990 & 0.688 & \textbf{0.841} \\
                                                            & & avgSIF & 0.992 & 0.750 & 0.992 & 0.670 & 0.988 & 0.670 & 0.831 \\
\cline{2-10}
                           & \multirow{2}{*}{$\mathcal{M}$-4} & adaSIF & 0.998 & \textbf{0.679} & 0.998 & 0.528 & 0.996 & 0.528 & \textbf{0.763} \\
                                                            & & avgSIF & 0.998 & 0.659 & 0.998 & 0.484 & 0.995 & 0.484 & 0.741 \\
\cline{2-10}
                           & \multirow{2}{*}{$\mathcal{M}$-5} & adaSIF & 0.985 & \textbf{0.625} & 0.985 & 0.408 & 0.965 & 0.408 & \textbf{0.697} \\
                                                            & & avgSIF & 0.994 & 0.608 & 0.994 & 0.359 & 0.982 & 0.359 & 0.676 \\
\cline{2-10}
                           & \multirow{2}{*}{$\mathcal{M}$-6} & adaSIF & 0.992 & \textbf{0.606} & 0.992 & 0.354 & 0.977 & 0.354 & \textbf{0.673} \\
                                                            & & avgSIF & 0.997 & 0.600 & 0.997 & 0.334 & 0.992 & 0.334 & 0.666 \\
\cline{2-10}
                           & \multirow{2}{*}{$\mathcal{M}$-7} & adaSIF & 0.982 & 0.599 & 0.982 & 0.342 & 0.951 & 0.342 & 0.662 \\
                                                            & & avgSIF & 0.995 & \textbf{0.601} & 0.995 & 0.339 & 0.986 & 0.339 & \textbf{0.667} \\
\hline
\hline
\multirow{14}{*}{CIFAR-100}& \multirow{2}{*}{$\mathcal{M}$-1} & adaSIF & 1.000 & \textbf{1.000} & 1.000 & 1.000 & 1.000 & 1.000 & \textbf{1.000} \\
                                                            & & avgSIF & 1.000 & \textbf{1.000} & 1.000 & 1.000 & 1.000 & 1.000 & \textbf{1.000} \\
\cline{2-10}
                           & \multirow{2}{*}{$\mathcal{M}$-2} & adaSIF & 0.990 & \textbf{0.994} & 0.990 & 0.994 & 0.990 & 0.994 & 0.992 \\
                                                            & & avgSIF & 0.996 & \textbf{0.994} & 0.996 & 0.994 & 0.996 & 0.994 & \textbf{0.995} \\
\cline{2-10}
                           & \multirow{2}{*}{$\mathcal{M}$-3} & adaSIF & 0.995 & \textbf{0.960} & 0.995 & 0.958 & 0.995 & 0.958 & \textbf{0.977} \\
                                                            & & avgSIF & 0.995 & 0.957 & 0.995 & 0.955 & 0.995 & 0.955 & 0.975 \\
\cline{2-10}
                           & \multirow{2}{*}{$\mathcal{M}$-4} & adaSIF & 0.994 & \textbf{0.883} & 0.994 & 0.868 & 0.993 & 0.868 & \textbf{0.931} \\
                                                            & & avgSIF & 0.988 & 0.866 & 0.988 & 0.848 & 0.986 & 0.848 & 0.918 \\
\cline{2-10}
                           & \multirow{2}{*}{$\mathcal{M}$-5} & adaSIF & 0.997 & \textbf{0.838} & 0.997 & 0.807 & 0.997 & 0.807 & \textbf{0.902} \\
                                                            & & avgSIF & 0.998 & 0.831 & 0.998 & 0.797 & 0.997 & 0.797 & 0.897 \\
\cline{2-10}
                           & \multirow{2}{*}{$\mathcal{M}$-6} & adaSIF & 0.986 & \textbf{0.837} & 0.986 & 0.808 & 0.983 & 0.808 & \textbf{0.897} \\
                                                            & & avgSIF & 0.990 & 0.795 & 0.990 & 0.744 & 0.987 & 0.744 & 0.867 \\
\cline{2-10}
                           & \multirow{2}{*}{$\mathcal{M}$-7} & adaSIF & 0.986 & \textbf{0.839} & 0.986 & 0.811 & 0.983 & 0.811 & \textbf{0.898} \\
                                                            & & avgSIF & 0.999 & 0.812 & 0.999 & 0.769 & 0.999 & 0.769 & 0.884 \\
\hline
\hline
\multirow{14}{*}{\parbox{1.5cm}{~~~~Tiny\\ImageNet}} & \multirow{2}{*}{$\mathcal{M}$-2} & adaSIF & 0.996 & 0.996 & 0.996 & 0.996 & 0.996 & 0.996 & 0.996 \\
                                                                                      & & avgSIF & 0.994 & \textbf{1.000} & 0.994 & 1.000 & 0.994 & 1.000 & \textbf{0.997} \\
\cline{2-10}
                         & \multirow{2}{*}{$\mathcal{M}$-3} & adaSIF & 0.998 & 0.989 & 0.998 & 0.988 & 0.998 & 0.988 & \textbf{0.993} \\
                                                            & & avgSIF & 0.989 & \textbf{0.990} & 0.989 & 0.990 & 0.989 & 0.990 & 0.990 \\
\cline{2-10}
                         & \multirow{2}{*}{$\mathcal{M}$-4} & adaSIF & 0.994 & \textbf{0.974} & 0.994 & 0.973 & 0.994 & 0.973 & \textbf{0.984} \\
                                                            & & avgSIF & 0.999 & 0.964 & 0.999 & 0.963 & 0.999 & 0.963 & 0.981 \\
\cline{2-10}
                         & \multirow{2}{*}{$\mathcal{M}$-5} & adaSIF & 0.998 & \textbf{0.960} & 0.998 & 0.958 & 0.998 & 0.958 & \textbf{0.978} \\
                                                            & & avgSIF & 0.997 & 0.956 & 0.997 & 0.954 & 0.997 & 0.954 & 0.976 \\
\cline{2-10}
                         & \multirow{2}{*}{$\mathcal{M}$-6} & adaSIF & 0.999 & \textbf{0.941} & 0.999 & 0.937 & 0.999 & 0.937 & \textbf{0.968} \\
                                                            & & avgSIF & 0.998 & 0.934 & 0.998 & 0.929 & 0.997 & 0.929 & 0.963 \\
\cline{2-10}
                         & \multirow{2}{*}{$\mathcal{M}$-7} & adaSIF & 0.997 & \textbf{0.935} & 0.997 & 0.931 & 0.997 & 0.931 & \textbf{0.964} \\
                                                            & & avgSIF & 0.992 & 0.930 & 0.992 & 0.926 & 0.991 & 0.926 & 0.959 \\
\bottomrule
\end{tabular}
}
\label{tab:adasif_vs_avgsif}
\vspace{-0.18in}
\end{table}
\clearpage
\section{Comparison of MI attacks with data augmentation}
\label{supp_sec:comparison_of_mi_attacks_with_data_augmentation}
\setcounter{figure}{0}
\renewcommand{\thefigure}{H\arabic{figure}}
Here we continue the adaptive MI attack comparison from Section~\ref{sec:data_augmentation_adaptive_attack}, and include other architectures. Figure~\ref{fig:alexnet_densenet_adaptive_attack_scores_small} presents the balanced accuracy on target models trained on AlexNet and DenseNet, with data augmentations (random crop and horizontal flipping). We observe that SIF performs on par with current SOTA, however, utilizing adaSIF (red bar) achieves new SOTA in most cases.

\begin{figure}[ht!]
\centering
\includegraphics[width=1\linewidth]{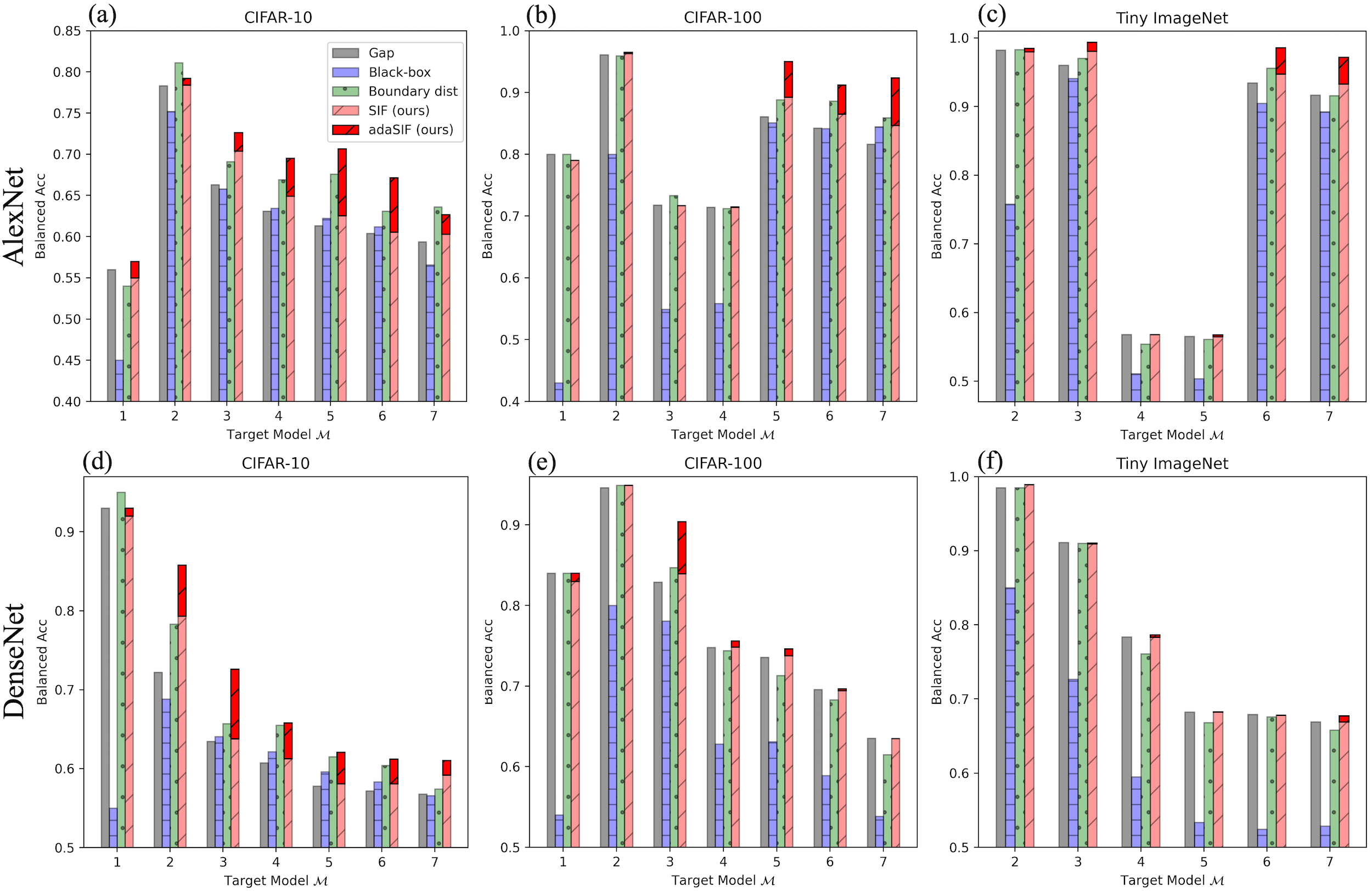}
\vspace{-0.24in}
\caption{Comparison of our SIF (pink bar) and adaSIF (red bar) attacks with some baseline MI attacks: Gap, Black-box, and Boundary distance. The top and bottom rows show target models trained with data augmentations on AlexNet and DenseNet architectures, respectively. The x-axis indicates the attacked target model and the y-axis shows the balanced attack accuracy (Eq.~\eqref{eq:balanced_accuracy}).}
\label{fig:alexnet_densenet_adaptive_attack_scores_small}
\vspace{-0.18in}
\end{figure}
\clearpage
\section{Comparison to a white-box attack}
\label{supp_sec:comparison_to_a_white_box_attack}
\setcounter{table}{0}
\renewcommand{\thetable}{I\arabic{table}}
Throughout our paper, we compare SIF and adaSIF to other SOTA black-box MI attacks, since researchers found that they perform similarly to white-box attacks \cite{Sablayrolles2019WhiteboxVB,Rezaei2021OnTD,leino2020stolen}. However, Nasr et al. presented higher balanced accuracy for their white-box attack compared to other black-box methods, by training a large DNN attack model which gets as input all the hidden activations and gradients along the target model's layers \cite{Nasr2018ComprehensivePA}. Since they did not publish a code, we compare our MI attacks to their reported performances on CIFAR-100 for the same pre-trained target models they used: AlexNet, ResNet110, and DenseNet\footnote{We utilized AlexNet, and DenseNet pre-trained DNNs from \href{https://github.com/bearpaw/pytorch-classification}{https://github.com/bearpaw/pytorch-classification}, which is the same repository that was used in \cite{Nasr2018ComprehensivePA} for getting pre-trained models. CIFAR-100 was trained on ResNet110 using the script in \href{https://github.com/bearpaw/pytorch-classification/blob/master/TRAINING.md}{https://github.com/bearpaw/pytorch-classification/blob/master/TRAINING.md} since its pre-trained weights could not be loaded on the updated architecture.}.

Table~\ref{tab:cifar100_ref} compares the balanced accuracy of different MI attacks on CIFAR-100 for the pre-trained models used in \cite{Nasr2018ComprehensivePA}. We show that our adaSIF attack achieves a new SOTA for all the pre-trained networks, outperforming the white-box attack of Nasr et al..
We emphasize that we trained ResNet110 for our experiments since the pre-trained ResNet110 weights in the repository that \cite{Nasr2018ComprehensivePA} relied on cannot be used anymore. Our ResNet110 train/test accuracies are $99\%/71\%$, whereas Nasr et al. used a model with train/test accuracies of $89\%/73\%$. This might explain the large gap in MI performance for ResNet110 between their method and adaSIF. 

We point out that our attack model utilizes only two fitted parameters ($\tau_1, \tau_2$), while Nasr et al. trained a heavy DNN for their attack model; this makes our method much more favorable for MI attack.

\begin{table*}[ht!]
\caption{Comparison between our SIF/avgSIF/adaSIF MI attacks and the white-box attack proposed by Nasr et al.\cite{Nasr2018ComprehensivePA}. For completeness, we report also the balanced accuracies of the black-box methods we used in the paper.}
\centering
\begin{tabular}{cc|c}
\toprule
Architecture & Attack model & Balanced Acc \\
\hline
\multirow{7}{*}{AlexNet}   & Gap             & 0.7421 \\
                           & Black-box       & 0.6549 \\
                           & Boundary dist   & 0.7362 \\
                           & Nasr et al.     & 0.7510 \\
\cdashline{2-3}
                           & SIF             & 0.7454 \\
                           & avgSIF          & \textbf{0.7594} \\
                           & adaSIF          & 0.7516 \\
\hline
\multirow{7}{*}{ResNet110} & Gap             & 0.6450 \\
                           & Black-box       & 0.6640 \\
                           & Boundary dist   & 0.6680 \\
                           & Nasr et al.     & 0.6430 \\
\cdashline{2-3}
                           & SIF             & 0.6616 \\
                           & avgSIF          & 0.6906 \\
                           & adaSIF          & \textbf{0.6944} \\
\hline
\multirow{7}{*}{DenseNet}  & Gap             & 0.5885 \\
                           & Black-box       & 0.7019 \\
                           & Boundary dist   & 0.5380 \\
                           & Nasr et al.     & 0.7430 \\
\cdashline{2-3}
                           & SIF             & 0.7242 \\
                           & avgSIF          & 0.7402 \\
                           & adaSIF          & \textbf{0.7474} \\
\bottomrule
\end{tabular}
\label{tab:cifar100_ref}
\end{table*}
\end{document}